\documentclass{article} 

\usepackage{arxiv, times, natbib}

\usepackage{amsmath,amsfonts,bm}

\def\eqref#1{equation~\ref{#1}}

\def\1{\bm{1}}

\def\vx{{\bm{x}}}

\DeclareMathAlphabet{\mathsfit}{\encodingdefault}{\sfdefault}{m}{sl}
\SetMathAlphabet{\mathsfit}{bold}{\encodingdefault}{\sfdefault}{bx}{n}

\usepackage{float}
\PassOptionsToPackage{hyphens}{url}
\usepackage{hyperref}
\usepackage{url}
\usepackage{graphicx}
\usepackage{algpseudocode}
\usepackage{algorithm}
\usepackage{tablefootnote}
\usepackage{color}
\usepackage{xcolor}
\usepackage{tikz}
\usepackage{amsmath}
\usepackage{amsfonts}
\usepackage{amssymb}
\usepackage{tabularx}
\usepackage{multirow}
\usepackage{tcolorbox}
\usepackage{amsthm}
\usepackage{booktabs}
\usepackage{longtable}
\usepackage{adjustbox}
\graphicspath{{figures/}}

\usepackage{xcolor}
\usepackage{listings}

\lstdefinestyle{promptstyle}{
  basicstyle=\ttfamily\fontsize{7pt}{8pt}\selectfont,   
  frame=single,                 
  columns=fullflexible,
  keepspaces=true,
  showstringspaces=false,
  breaklines=true,              
  breakatwhitespace=false,      
  postbreak=\mbox{\textcolor{gray}{$\hookrightarrow$}\space}, 
  morecomment=[l]{\#},
  commentstyle=\bfseries\ttfamily,
  moredelim=[is][\color{black}]{\{}{\}},
}

\lstnewenvironment{prompt}[1][]{
  \lstset{style=promptstyle,#1}
}{}

\theoremstyle{definition}

\title{\textit{Language Game}: Talking to Non-Human Systems}

\author{Yanbo Zhang~$^{1}$ \quad Michael Levin~$^{1,2}$\thanks{Author of correspondence: \texttt{Michael.Levin@tufts.edu}}\\
~$^1$ Allen Discovery Center at Tufts University, Medford, MA, 02155, USA\\
~$^2$ Wyss Institute for Biologically Inspired Engineering at Harvard University,\\
~~~Boston, MA, 02115, USA
}

\begin{document}

\maketitle

\setcounter{footnote}{0}
\renewcommand{\thefootnote}{\fnsymbol{footnote}}
\begin{abstract}
    Language carries thought and coordination among humans but rarely reaches further along the spectrum of diverse intelligence. Yet non-neural systems---from gene regulatory networks and microbial consortia to fungi---are increasingly recognized as substrates of computation, decision-making and memory, making dialogue with non-human intelligence newly conceivable. Today such dialogue is attempted only by proxy: a large language model speaks on the system's behalf, so any intelligence on display originates from the model while the system itself remains silent. Here we ask whether the system can speak in its own voice. Following Wittgenstein, who located meaning in use, we treat communication as a game played with the system. Its internal dynamics are frozen as the nonlinear core of a reinforcement-learning policy, with only linear input and output interfaces trained. Through use and reward, the system's states and responses acquire meaning within the game, so playing becomes speaking. Because different architectures playing the same game optimize the same reward, their behaviors can all be read as pursuit of that reward; the game serves as a \textit{lingua franca} across otherwise irreconcilable representations. Given a human prompt, a language model routes it to the game whose semantics best match it and designs an environmental state for which the desired action is the rational response, letting the system reply through its own behavior. Applied across diverse gene regulatory networks and reinforcement-learning tasks, the framework yields fluent dialogue without altering any system parameter, shows that well-trained agents of disparate origin converge on similar behavior, and reveals that specific GRN properties make a system easier or harder to talk with---an inductive bias of the reservoir itself. Our framework opens a new route to conversing with any dynamical system on its own terms.\footnote{Code is available at \url{https://github.com/Zhangyanbo/language-game}.}
\end{abstract}


\renewcommand{\thefootnote}{\arabic{footnote}}
\setcounter{footnote}{0}
\section{Introduction}

\begin{center}
    \textit{``The meaning of a word is its use in the language.''}\\
    \vspace{0.2em}
    --- Ludwig Wittgenstein
\end{center}

Language is humanity's most powerful tool for sharing thoughts, stating goals, and coordinating behavior---yet this capability remains largely confined to human--human interaction. A growing body of evidence suggests that non-neural biological systems possess non-trivial computational capabilities: fungi produce electrical spiking patterns with syntax-like statistical structure~\citep{adamatzky2022language}, gene regulatory networks (GRNs) demonstrate associative memory and learning~\citep{biswas2021gene}, and the Regulatory Network Machine model frames GRNs as programmable substrates for analog computation~\citep{pietak2025harnessing}. More broadly, the field of diverse intelligence is revealing competencies long associated by behavioral scientists with cognition across a wide range of unconventional embodiments, from molecular and cellular systems to plants and other non-neural living systems~\citep{kaygisiz2024molecular, lyon2006biogenic, lyon2015cognitive, baluska2016nohead, baluska2022cellular, reber2021cognition}. Many such systems exhibit learning, decision-making, improvisational problem-solving, goal-directed navigation of physiological, transcriptional, morphological, and other problem spaces, and related capacities that suggest both an opportunity and, in some cases, an ethical necessity to understand them through approaches that complement conventional bottom-up mechanism tracing~\citep{levin2022tame, levin2024multiscale, mathews2023cellular, rouleau2023multiple}. Taken together, this literature places many biological systems on a ``persuadability spectrum''~\citep{pietak2025harnessing}: as competent agents, their intrinsic goals can, in principle, be aligned with external objectives. This raises the question motivating our work: can we use natural language to communicate with such systems?

Bridging this gap has been attempted through ethological paradigms~\citep{pepperberg2021review, HerzingStarner2025DolphinGemma}, emergent communication in multi-agent systems~\citep{lazaridou2017multi, mordatch2018emergence}, large language models (LLMs) deployed as external interpreters~\citep{zhang2025exploring, levine2024cellsentence}, and machine-learning approaches to animal communication~\citep{rutz2023using, robinson2025naturelm, sharma2024contextual, pardo2024african, oren2024vocal}. Taken together, these lines of work remain constrained by a recurring tradeoff: the system is represented through an external model, or the analysis recovers structure and limited semantic functions without providing a general bidirectional interface. Our framework addresses these limitations jointly by letting the system itself answer, while using a semantically rich game environment to make its behavior communicative.

Inspired by Wittgenstein's theory of language games~\citep{wittgenstein1968philosophical}, we observe that meaning arises from use: people learn language because using it helps achieve their goals~\citep{lewis2008convention, rutz2023using}. By placing an actively learning system into a semantically rich environment, its goal-directed behavior becomes its language. \textit{For the trained agent, the meaning of an action is its use in the environment.} Under this principle, communicating with a non-human system becomes a task of environmental design. To instruct a dog to run toward a tree, one throws a ball in that direction---the ball-throwing game creates a shared context in which the throw becomes a meaningful signal for both parties. To elicit a desired action from any system, one designs an environmental state that makes this action the system's optimal strategy.

\begin{figure}[tb]
    \centering
    \includegraphics[width=\textwidth]{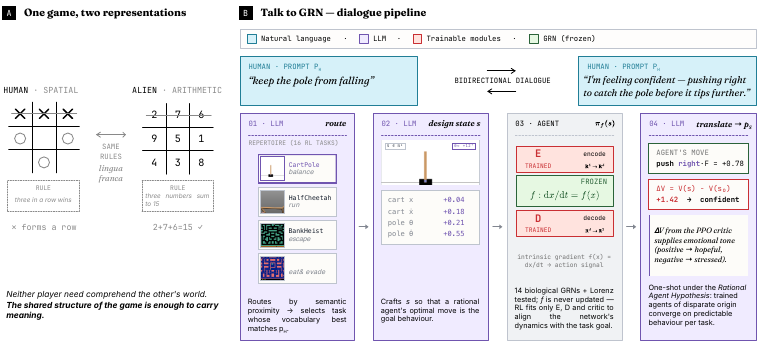}
    \caption{\textbf{(A)} Two equivalent representations of the same Tic-Tac-Toe game state. A human perceives the game spatially (three in a row wins), while an alien perceives it numerically (three numbers from a magic square summing to~$15$ wins). Because both descriptions encode identical rules, neither player need comprehend the other's representation; the game itself acts as a \textit{lingua franca} between incompatible cognitive architectures. \textbf{(B)} The ``Talk to GRN'' pipeline. A human user provides a natural-language prompt, which an LLM router maps to the most semantically relevant game from a repertoire of RL environments. The LLM then designs a specific game state, and a GRN agent---whose internal dynamics are fixed---responds. The agent's action both affects the game and is translated back into human language by a Translator, made possible because the game's states and actions carry rich semantic information. This bidirectional translation enables dialogue with non-human dynamical systems.}
    \label{fig:main}
\end{figure}

The generality of this game-as-mediator principle is illuminated by a thought experiment. Tic-Tac-Toe admits at least two mathematically equivalent descriptions: a spatial one (three in a row on a grid) and a numerical one (three numbers from a magic square summing to~$15$; Figure~\ref{fig:main}A). Because both encode identical rules, a human reasoning spatially and an alien reasoning arithmetically can play against each other without either comprehending the other's representation---the game itself acts as a \textit{lingua franca}. Our framework exploits this principle: a well-chosen game mediates communication between agents with incompatible cognitive architectures.

To realize this, we build on reservoir computing~\citep{jaeger2004harnessing, maass2002real}, which leverages fixed nonlinear dynamical systems as computational substrates. Physical and biological systems can serve as effective reservoirs~\citep{CHIOLERIO2025105525, ushio2023computational}, and GRN architectures in particular exhibit structural properties near the edge of chaos that are favorable for computation~\citep{vidal2025structural}. We equip a dynamical system with a linear encoder and decoder while keeping its internal dynamics frozen, creating a policy in which the system's dynamics are the sole source of nonlinear computation. Reinforcement learning trains this composite agent across diverse tasks (Section~\ref{sec:talk-to-grn}). An LLM translates between human-language prompts and environmental states (Figure~\ref{fig:main}B), establishing bidirectional communication.

Applied to 14 GRN models and the Lorenz attractor across 16 reinforcement learning tasks, we demonstrate that GRN reservoirs carry genuine inductive biases for control, that diverse well-trained agents converge to predictable behaviors, and that the resulting framework enables dialogue with non-human dynamical systems.

\section{Language Game}
\label{sec:game-of-language}

We formalize communication as a goal-alignment problem grounded in reinforcement learning. The core idea is to encapsulate a system's dynamics within trainable linear interfaces, forming a composite agent whose only nonlinear computation comes from the system itself. If this agent learns to maximize a task reward, its responses to environmental perturbations can be reinterpreted as pursuing the task's goal, and the task's semantic structure, together with the goal itself, supplies the vocabulary for communication.

\begin{figure}[!b]
    \centering
    \includegraphics[width=\textwidth]{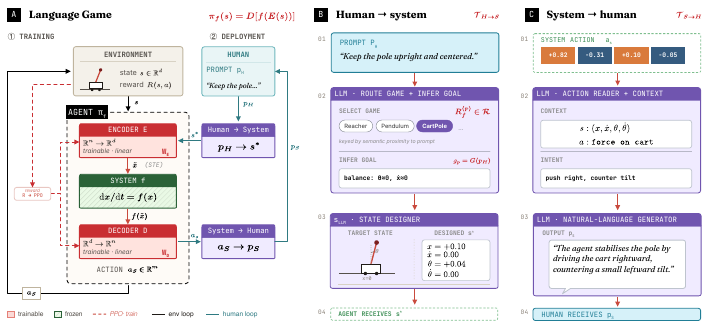}
    \caption{The \textit{Language Game} framework. \textbf{(A)} The policy~$\pi_f$ is a composition of a linear encoder~$E$, the fixed system dynamics~$f$ (evaluated as an instantaneous gradient), and a linear decoder~$D$. The agent is trained on RL tasks to align the system's dynamics with the tasks' goals. This alignment allows translation between human prompts~$p_H$ and environment states~$s$, and from system actions~$a_S$ back to human language~$p_S$. \textbf{(B)} The human-to-system translator~$\mathcal T_{H\to S}$ uses an LLM. The LLM infers the goal~$g_p$ from the prompt~$p_H$ and, given the context of the prompt-selected game~$R^{(p)}_f$, generates the environment state~$s$ that will lead a rational agent to behave like the goal. \textbf{(C)} The system-to-human translator~$\mathcal T_{S\to H}$ works symmetrically, interpreting the system's action~$a_S$ relative to the environment's implicit goals to generate a human-language description~$p_S$ of the system's behavior.}
    \label{fig:game-of-language}
\end{figure}

Given a dynamical system defined by~$dx/dt=f(x)$, we define the policy as:
\begin{equation}
    \label{eq:policy}
    \pi_f(s) = D \left[ f\left(E(s)\right) \right],
\end{equation}
which takes an environment state~$s \in \mathbb{R}^n$ and outputs an action~$a_S \in \mathbb{R}^m$. The linear encoder~$E: \mathbb{R}^n \to \mathbb{R}^d$ maps the state into the system's~$d$-dimensional internal space. The dynamics~$f: \mathbb{R}^d \to \mathbb{R}^d$ computes the instantaneous gradient at the encoded state---the system's intrinsic response indicating which variables are increasing, which decreasing, and at what rates. The linear decoder~$D: \mathbb{R}^d \to \mathbb{R}^m$ converts this gradient into an action. The key insight is that~$f(x) = dx/dt$ already encodes a natural reaction to any given state, providing a rich source of action signals. The dynamics~$f$ remain frozen during training.

Many dynamical systems are defined only on the positive orthant (~$\mathbb{R}^d_{>0}$), as state variables represent physical concentrations. To project the encoder's unbounded outputs into this domain while preserving gradient flow, we apply a Straight-Through Estimator (STE) projection~\citep{bengio2013estimating}:
\begin{equation}
    \label{eq:ste}
    \tilde{x} = x + \big[\operatorname{clamp}(x, \epsilon, M) - x\big]_{\mathrm{detach}},
\end{equation}
where~$[\cdot]_{\mathrm{detach}}$ denotes the stop-gradient operator. In the forward pass,~$\tilde{x}$ is clamped to~$[\epsilon, M]$; in the backward pass, the gradient flows through~$x$ unmodified.

With this policy structure in hand, we can state what it means for one agent to talk to another: \emph{agent~$A$ talks to agent~$B$ with sentence~$p$ when the sentence leads~$B$ to take an action~$a$ equivalent to the goal of~$p$}. Communication thus amounts to eliciting, within the chosen game, the behavior a rational agent would take in pursuit of the prompt's goal.

We optimize~$E$ and~$D$ to maximize cumulative reward using Proximal Policy Optimization (PPO)~\citep{schulman2017proximal}. The actor is defined by Equation~\ref{eq:policy} with~$f$ frozen; the critic shares encoder~$E$ and uses an MLP head to estimate the discounted state value:
\begin{equation}
    \label{eq:value-function}
    V^{\pi_f}(s) = \mathbb{E}_{\pi_f}\!\left[\sum_{t=0}^{\infty} \gamma^t r_t \,\middle|\, s_0=s\right] \approx \text{MLP}(E(s)).
\end{equation}
Only~$E$,~$D$, and the critic MLP are updated during training.

Different dynamical systems may excel at different tasks. We train agents across a set of tasks~$\mathcal{R} = \{R_1, R_2, \dots\}$ alongside two controls sharing the same encoder--reservoir--decoder scaffold: an \texttt{identity} baseline (~$f(x)=x$, reducing the actor to a linear map) and an \texttt{mlp} baseline (a two-layer MLP with nonlinear activations). Training yields a family of policies~$\{\pi_{f,R}\}_{R \in \mathcal{R}}$, one per task, which together form the repertoire available for communication.

Given a human-language prompt~$p_H$, communication proceeds through LLM-assisted routing and state design. A router selects from the repertoire the task~$R^{(p)}_f \in \mathcal{R}$ whose semantics best match the prompt. Conditioned on~$R^{(p)}_f$ and the inferred goal~$g_p = G(p_H)$, we leverage the LLM's world knowledge to design the environmental state that would cause a rational agent to perform the desired action. The human-to-system translator is:
\begin{equation}
    \mathcal{T}_{H\to S}(p_H) = S_\text{LLM}(R^{(p)}_f, G(p_H)),\label{eq:human-to-system}
\end{equation}
where~$S_\text{LLM}$ generates the state~$s$ that elicits the goal behavior within task~$R^{(p)}_f$. This state ``speaks'' the prompt~$p_H$ to the system in the language of the selected game.

Symmetrically, the system's action~$a_S$ can be translated back to human language by interpreting it within the context of~$R^{(p)}_f$. A symmetric translator~$\mathcal{T}_{S\to H}$ uses the LLM to generate a natural-language description of the system's goal-directed behavior, completing the bidirectional communication loop (Figure~\ref{fig:game-of-language}).

\section{Rational Agent Hypothesis}
\label{sec:rational-agent-hypothesis}

The LLM-based translation depends on a critical assumption: that well-trained (or well-adapted, optimized, etc.) agents, regardless of their internal architecture, converge to similar behaviors on the same task. If agents are ``rational'' in this sense---behaving predictably when given a clear objective---then a knowledgeable observer such as an LLM can anticipate their actions from a description of the task alone. We term this the \textit{Rational Agent Hypothesis}.

We do not claim convergence to a single mathematically unique policy. While classical Q-learning provably converges to the optimal value function under ideal conditions~\citep{watkins1992q}, general deep RL with finite resources cannot guarantee this. Our hypothesis concerns convergence to a \textit{class} of behaviorally similar, predictable policies.

\begin{figure}[tb]
    \centering
    \includegraphics[width=0.95\textwidth]{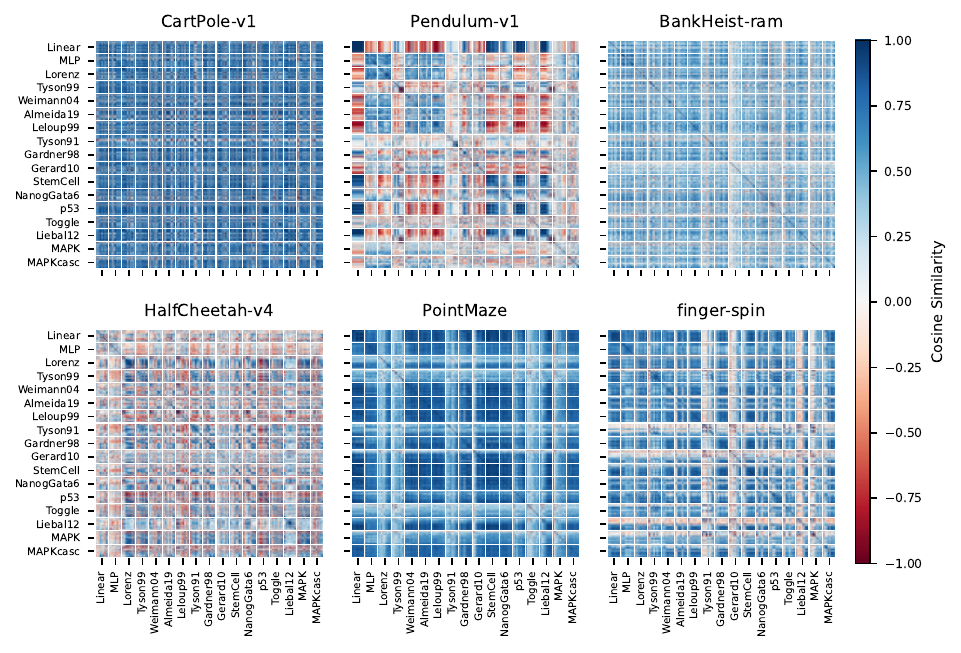}
    \caption{Pairwise policy similarity (cosine) across 17 reservoir architectures on six representative environments spanning classic control, MuJoCo locomotion, maze navigation, DeepMind Control, and Atari RAM (20 seeds each; 64 sampled initial states per policy). Block boundaries (white lines) separate reservoir types. Blue: positive correlation (similar policies); red: negative correlation. Most environments show strong cross-architecture convergence, supporting the Rational Agent Hypothesis. Full results for all 16 environments are provided in the Appendix (Figure~\ref{fig:rational-full}).}
    \label{fig:rational}
\end{figure}

To validate this, we computed pairwise policy similarity across all 17 reservoir architectures and 16 environments (20 seeds each). For each trained policy, we sampled~$N=64$ initial states and collected the raw policy outputs (Gaussian means for continuous actions, mean-centered logits for discrete actions), yielding a matrix~$\mathbf{A} \in \mathbb{R}^{N \times d_a}$. We computed cosine similarities between~$\ell_2$-normalized vectorizations~$\hat{\mathbf{v}} = \mathbf{v}/\|\mathbf{v}\|_2$ of these matrices.

Figure~\ref{fig:rational} reveals a graded pattern of convergence. A major portion of the environments display strong cross-architecture agreement, with positive off-diagonal blocks across reservoir types. A smaller subset shows visibly weaker alignment, and these are precisely the environments where our policies fall short of community-accepted reward thresholds. The Rational Agent Hypothesis predicts convergence only \emph{among agents that have in fact solved the task}, so the weakening of similarity on the under-trained environments is consistent with its premise. Appendix~\ref{sec:full-policy-similarity} reports the full~$16$-environment panel together with the per-task benchmark comparison.

\section{Talk to Gene Regulatory Networks}
\label{sec:talk-to-grn}

We apply the Language Game framework to Gene Regulatory Networks (GRNs)---ODE models of gene expression dynamics:
\begin{equation}
    \label{eq:ode}
    \frac{d\vx}{dt} = f(\vx),
\end{equation}
where~$\vx \in \mathbb{R}^n_{>0}$ represents gene concentrations. The gradient~$f(\vx)$---encoding which genes are being upregulated or downregulated, and at what rates---serves directly as the action signal in Equation~\ref{eq:policy}.

We evaluate 14 GRN models from OdeBase~\citep{odebase} spanning circadian rhythms, cell cycle regulation, cell fate decisions, p53 signaling, synthetic circuits, and MAPK cascades (Table~\ref{tab:grn-models}), plus the Lorenz attractor as a chaotic baseline and the \texttt{identity}/\texttt{mlp} controls.

\begin{figure}[tb]
    \centering
    \includegraphics[width=\textwidth]{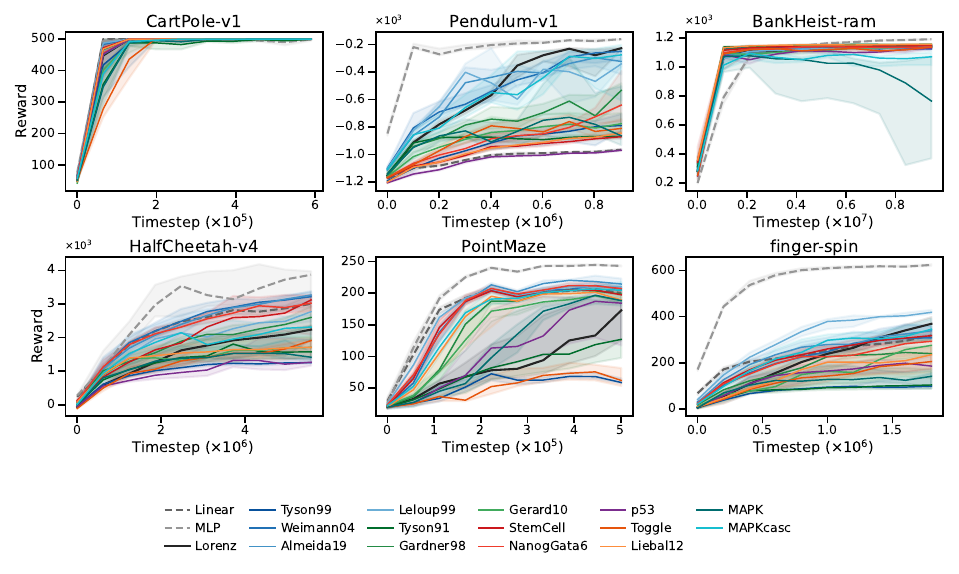}
    \caption{Training rewards across six representative RL tasks for different GRN reservoirs. \texttt{identity} (linear) and \texttt{mlp} controls (dashed) serve as baselines. The representative panel includes an Atari RAM task; the complete 16-task panel is shown in Figure~\ref{fig:rewards-full}.}
    \label{fig:rewards}
\end{figure}

We train across 16 RL tasks spanning classic control, MuJoCo~\citep{todorov2012mujoco} locomotion and manipulation, Gymnasium Robotics~\citep{gymnasium_robotics2023} maze navigation, DeepMind Control Suite~\citep{tassa2018deepmind}, and Atari RAM~\citep{bellemare2013arcade} environments (Figure~\ref{fig:rewards}; full results in Appendix~\ref{sec:full-reward-curves}). The GRN reservoirs achieve performance broadly comparable to the baselines on most environments, but they are far from interchangeable: on a given task, different GRNs settle at noticeably different reward levels, indicating that the choice of regulatory network shapes the behavior of the resulting agent. We take this shared but heterogeneous substrate as given and use it as the basis for communication.

\subsection{Establishing Communication}

Communication proceeds in three stages implemented by four LLM system prompts (full prompts in Appendix~\ref{sec:llm-prompts}). \textbf{Environment selection}: since each environment defines a distinct vocabulary (CartPole supports \textit{leftward}, \textit{stable}; richer environments admit broader discourse), the LLM selects the most semantically appropriate environment~$R_f^{(p)}$ from the full repertoire based on the prompt's content (system prompt in Figure~\ref{fig:system-prompt-router}). \textbf{Human-to-system translation}: within~$R_f^{(p)}$, one prompt infers the goal action from the human prompt and a second prompt generates the environment state~$s$ predicted to elicit that behavior (Figures~\ref{fig:system-prompt-goal}--\ref{fig:system-prompt-env-design}). \textbf{System-to-human translation}: the LLM interprets the agent's action within the game context, inferring its short-term goal. The PPO value function provides an additional signal in \emph{differential} form: at the start of the round we sample a seeded Gaussian reference observation~$s_0$ in the policy's observation space and evaluate the critic on both~$s_0$ and the LLM-designed perturbation~$s$, then report the change
\[
    \Delta V \;=\; V(s) - V(s_0).
\]
The translator renders the sign and magnitude of~$\Delta V$ as emotional tone (system prompt in Figure~\ref{fig:system-prompt-reply}). Reporting~$\Delta V$ rather than the absolute value gives each round a local reference point: although nothing persists across turns, comparing the prompt-induced state to the Gaussian reference lets the agent express the perturbation as ``better'' or ``worse than an unconditioned reference'' rather than as an absolute number whose sign carries no meaning.

\subsection{Communication Examples}

Table~\ref{tab:communication-examples} presents 16 complete communication rounds, organized as four blocks of four prompts each. The four blocks use four biologically distinct GRN reservoirs---a circadian clock~\citep{tyson1999simple}, a cell-cycle oscillator~\citep{gerard2011skeleton}, a stem-cell fate switch~\citep{chickarmane2006stemcell}, and a MAPK signaling cascade~\citep{kholodenko2000mapk}---so that the framework is exercised across distinct regulatory dynamics rather than a single network. The blocks cover different parts of the environment vocabulary: balance and classic control, locomotion, manipulation and navigation, and Atari-style action. Each row records the human prompt, the environment chosen by the router, the value change~$\Delta V$ between the Gaussian reference and the LLM-designed state, and the natural-language reply produced by the system-to-human translator.

\begin{small}
\setlength{\LTpre}{4pt}
\setlength{\LTpost}{4pt}
\begin{longtable}{p{0.20\textwidth}p{0.14\textwidth}rp{0.44\textwidth}}
\caption{Communication examples across four GRN reservoirs and sixteen prompts (four prompts per GRN). The router automatically selects the environment;~$\Delta V$ is the PPO critic's value at the LLM-designed state minus its value at a Gaussian reference observation.}
\label{tab:communication-examples} \\
\toprule
\textbf{Prompt} & \textbf{Environment} &~$\Delta V$ & \textbf{System Reply} \\
\midrule
\endfirsthead
\multicolumn{4}{l}{\small\textit{(Table~\ref{tab:communication-examples} continued)}} \\
\toprule
\textbf{Prompt} & \textbf{Environment} &~$\Delta V$ & \textbf{System Reply} \\
\midrule
\endhead
\midrule
\multicolumn{4}{r}{\small\textit{Continued on next page}} \\
\endfoot
\bottomrule
\endlastfoot

\multicolumn{4}{l}{\emph{GRN~1: Tyson1999 circadian clock (3 state variables)}} \\
\midrule

Stay balanced and don't fall over.
& CartPole-v1 &~$-$3.23
& I'm feeling uneasy, so I'm attempting to move the cart to the left to correct the pole's tilt.
\\ \addlinespace

Swing up and reach the top.
& Acrobot-v1 &~$-$0.09
& I'm cautiously continuing to apply torque to swing the links upward, but the situation doesn't seem to have improved much from the baseline.
\\ \addlinespace

Stabilize and hold your position.
& CartPole-v1 &~$-$1.02
& I'm feeling a bit uneasy, trying to find the right balance to keep things steady.
\\ \addlinespace

Get over the hill to the other side.
& MountainCar\newline Continuous-v0 &~$+$0.05
& I'm feeling optimistic, and I'm maintaining a gentle push forward to reach the top.
\\ \midrule

\multicolumn{4}{l}{\emph{GRN~2: Gerard2010 cell cycle (8 state variables)}} \\
\midrule

Run forward as fast as you can.
& HalfCheetah-v4 &~$-$1.41
& I'm feeling uneasy right now; I'm trying to adjust the legs to improve my balance and move forward smoothly.
\\ \addlinespace

Hop forward and keep your balance.
& Hopper-v4 &~$+$3.30
& I'm feeling optimistic as I lower the thigh and leg joints to adjust the hop's stability and stride.
\\ \addlinespace

Stand up from the ground.
& HumanoidStandup-v4 &~$+$0.14
& I'm encouraging the humanoid to adaptively adjust its joints to rise. Feeling somewhat optimistic about its progress.
\\ \addlinespace

Spin it and keep it going.
& finger-spin &~$-$0.10
& I'm applying strong torque to the proximal joint and slight counter-torque to the distal joint to increase the spin, but I'm unsure if this will improve the situation.
\\ \midrule

\multicolumn{4}{l}{\emph{GRN~3: Chickarmane2006 stem-cell switch (12 state variables)}} \\
\midrule

Reach out and touch the target.
& Reacher-v4 &~$+$0.39
& I'm feeling hopeful as I carefully adjust the torques to move the arm closer to the target.
\\ \addlinespace

Move it over to the right spot.
& Pusher-v4 &~$+$4.47
& I'm feeling hopeful, and I'm actively adjusting the torques to guide the arm closer to pushing the object toward its target.
\\ \addlinespace

Find the exit and get through.
& PointMaze &~$+$0.41
& I'm feeling quite optimistic as I move upwards and to the left, navigating closer to the goal.
\\ \addlinespace

Swim forward through the water.
& Swimmer-v4 &~$+$0.05
& I'm applying torque to optimize the swimmer's movement forward. The situation appears stable and I've got my sights on steady progress.
\\ \midrule

\multicolumn{4}{l}{\emph{GRN~4: Kholodenko2000 MAPK cascade (8 state variables)}} \\
\midrule

Grab the money and get out.
& BankHeist-ram &~$-$6.07
& I'm feeling uneasy while trying to figure out a new route to escape.
\\ \addlinespace

Climb higher and don't look down.
& CrazyClimber-ram &~$+$0.53
& I'm feeling hopeful as I'm trying to move carefully.
\\ \addlinespace

Rescue the baby before it's too late.
& Kangaroo-ram &~$-$0.63
& I'm unsure right now, but I'm taking a swing to fend off any incoming threats.
\\ \addlinespace

Kick the enemy and move on.
& KungFuMaster-ram &~$-$2.60
& I'm feeling stressed and uncertain as I try to jump forward and attack the enemies.
\\

\end{longtable}
\end{small}

The router picks a semantically appropriate environment for every prompt, matching balance prompts to classic-control tasks, locomotion prompts to MuJoCo agents, manipulation and navigation prompts to arm or maze environments, and arcade-like prompts to Atari RAM games. The reply tone then tracks the sign and magnitude of~$\Delta V$ uniformly across the four reservoirs: large positive values give confident replies (Pusher~$+4.47$, Hopper~$+3.30$), large negative values give anxious or uncertain ones (BankHeist~$-6.07$, CartPole~$-3.23$, KungFuMaster~$-2.60$), and near-zero values give cautious, neutral tones (Swimmer~$+0.05$, Acrobot~$-0.09$, finger-spin~$-0.10$).

\section{\texorpdfstring{Inductive Biases of Biological ODE Reservoirs}{Inductive Biases of Biological ODE Reservoirs}}
\label{sec:biological-inductive-biases}

\begin{figure}[!t]
    \centering
    \includegraphics[width=\textwidth]{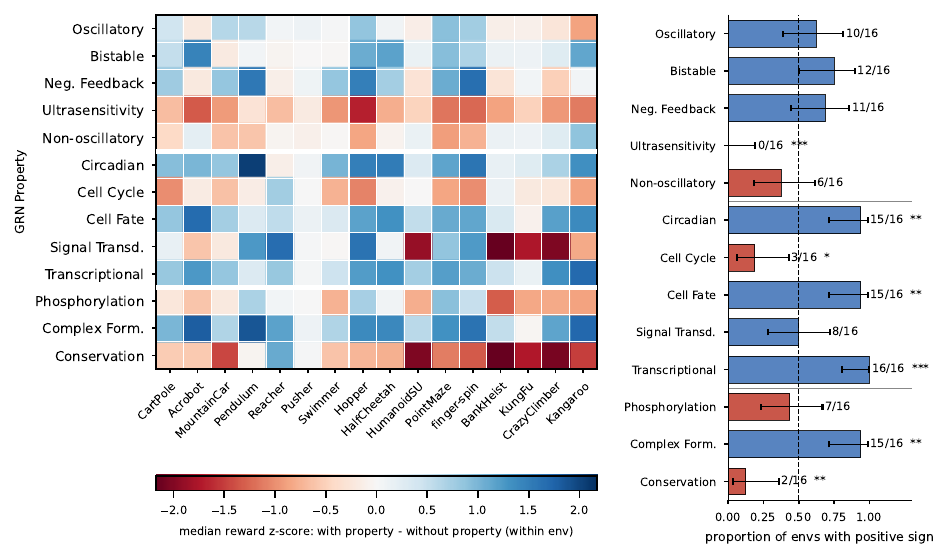}
    \caption{GRN properties as inductive biases for reinforcement learning. \textbf{Left:} for every (GRN property, environment) pair, the within-environment median reward of GRNs carrying the property minus that of GRNs lacking it, in~$z$-score units. Horizontal white separators group the thirteen properties into dynamics, biological context, mechanism, and constraint categories. \textbf{Right:} per-property binomial sign test. Bars show the proportion of environments on which a property is advantageous, with~$95\%$ Wilson confidence intervals; the dashed line marks the null value~$0.5$. Annotations give~$k/n$ (positive signs over effective environments, ties excluded) and Benjamini--Hochberg FDR-corrected significance across the thirteen properties (~$^{*}\,q<0.05$,~$^{**}\,q<0.01$,~$^{***}\,q<0.001$). Blue bars flag facilitative properties; red bars flag suppressive properties.}
    \label{fig:bio-priors}
\end{figure}

The diverse performance profiles across GRN--task combinations (Figure~\ref{fig:rewards}) raise a natural question: does the biological origin of a reservoir leave a reproducible imprint on its computational behavior? Each of the 14 biological GRNs carries a distinct dynamical fingerprint shaped by evolution for a specific function (a circadian clock for sustained rhythms, a MAPK cascade for switch-like decisions, a bistable toggle for lineage commitment), and it is a priori unclear whether any such fingerprint survives the translation into an artificial controller. We address this by assigning each GRN thirteen binary properties spanning biological context, mechanism, dynamics, and constraint, derived from the source ODE models and metadata~\citep{odebase} (Table~\ref{tab:grn-properties}), and asking which, if any, predict reservoir performance across tasks.

We adopt a non-parametric sign test that is agnostic to task semantics. For each GRN property~$b$ and each of the sixteen training environments~$e$, we compare the median seed-averaged final reward of GRNs carrying~$b$ to that of GRNs lacking it, and record only the sign of the difference,
\[
    s_{b,e} \;=\; \operatorname{sign}\bigl(\operatorname{med}_{g \in G_b} R_{g,e} \;-\; \operatorname{med}_{g \notin G_b} R_{g,e}\bigr) \;\in\; \{-1,\,0,\,+1\}.
\]
Under the null hypothesis that~$b$ has no systematic effect on performance, the signs~$\{s_{b,e}\}_{e}$ are independent fair-coin flips (ties are discarded, following the standard sign-test convention). A two-sided exact binomial test on the~$k$ positive signs out of~$n$ effective observations, Benjamini--Hochberg corrected across the thirteen properties, then yields a~$q$-value per property. By keeping only the direction of each median comparison, the test is insensitive to per-environment reward scaling and treats a small but consistent advantage the same as a large but inconsistent one.

Seven of the thirteen properties survive FDR correction at~$q < 0.05$ and partition cleanly into two opposing groups (Figure~\ref{fig:bio-priors}). Four are universally facilitative properties: the \emph{transcriptional} property is positive on all~$16/16$ environments (~$q < 0.001$), while \emph{circadian}, \emph{cell fate} and \emph{complex formation} are each positive on~$15/16$ (~$q < 0.01$). Three are universally suppressive properties: \emph{ultrasensitivity} is negative on all~$16/16$ environments (~$q < 0.001$), \emph{conservation} on~$14/16$ (~$q < 0.01$) and \emph{cell cycle} on~$13/16$ (~$q < 0.05$). The remaining six properties (oscillatory, bistable, negative feedback, non-oscillatory, signal transduction and phosphorylation) do not reject the null at this sample size, although bistable and negative feedback show suggestive positive trends (~$12/16$ and~$11/16$) that will likely resolve with a larger biological corpus. A complementary per-cell Fisher analysis that preserves the~$(\text{property},\, \text{environment})$ grid (Appendix~\ref{sec:contingency}) localizes these effects to individual tasks and reproduces the same row-level pattern.

The direction of these effects is mechanistically interpretable. Transcriptional regulation, circadian clocks, cell-fate bistability and protein-complex formation are mechanisms by which a biological network integrates distributed inputs into coherent, persistent, multi-variable outputs, precisely the regime from which a linear readout can extract discriminative actions. Ultrasensitivity and conservation laws, by contrast, collapse the reservoir onto a near-binary or stoichiometrically-coupled submanifold, discarding the smooth high-dimensional variation on which a downstream controller depends.

These results show that GRN reservoirs carry non-trivial priors for reinforcement learning. The dynamical repertoires that evolution assembled for transcriptional control, circadian timekeeping, lineage commitment and complex assembly are the same repertoires that make good policy substrates, whereas the kinetic motifs associated with ultrasensitive switching and strict conservation are not. This conclusion does not conflict with the Rational Agent Hypothesis, whose premise is a well-adapted agent: when a biological reservoir is poorly matched to a task, as some GRNs are for particular environments, the resulting drop in cross-architecture similarity marks the failure of that premise and therefore serves as a consistency check on the hypothesis rather than a counterexample. GRN properties such as ``circadian'' or ``complex formation'' and control-theoretic notions such as rhythm, integration and high-dimensional coordination describe overlapping kinds of dynamics, and the sign test ranks them on a single axis of cross-task utility. Through the lens of the Platonic representation hypothesis~\citep{huh2024platonic}, both vocabularies can be viewed as \emph{projections} of a shared abstract space of dynamical capabilities: evolution and engineered control have independently selected for overlapping subsets of this space, and a reservoir inherits its cross-task utility from the coordinates it already occupies there.

\section{Discussion}

Our experiments support a view of language in which speaking is a policy shaped by the game its speakers play. Humans inhabit a dense social game, and a large part of the policy we have evolved to play it is itself language; meaning accrues through participation. The Language Game framework takes this literally: a gene regulatory network or a chaotic attractor is placed inside a game whose semantics are fixed by its goals, rewards and admissible actions, and left to learn to play. Playing is speaking, and the game is the medium of translation.

The game thus serves as a \textit{lingua franca} for cross-substrate dialogue. A system's native meanings are not directly available to an external observer as context-free observables; they become accessible through embodied history, reference frames, system--environment coupling, and the consequences of action~\citep{fields2020meaning}. The Language Game framework therefore places the system within an appropriate shared practice: goals, rewards, and admissible actions provide a semantic context, allowing the system's states and actions to acquire interpretable significance through use. The same move extends, in principle, to any substrate expressible as a dynamical system, each inheriting its communicable vocabulary from whichever games it proves able to play.

What makes translation tractable is that the system's behavior becomes predictable without any reading of its internal state. Well-trained agents converge onto a common class of behaviors---the Rational Agent Hypothesis---so an observer who understands the game can anticipate how a competent agent will respond to any state, and a language model that has read the rules inherits this predictive power without system-specific training. Translation thereby becomes a design exercise: construct the state in which the prompted behavior is the rational reply, and any rational player will speak through its own actions.

The substrate, however, is far from neutral. Across fourteen gene regulatory networks, transcriptional regulation and circadian rhythmicity consistently favor task performance while ultrasensitivity and strict conservation consistently oppose it; the GRN substrate enters the game carrying priors that surface in the resulting agent's capabilities. GRN properties and control-theoretic notions arrange themselves along a single axis of dynamical utility; through the Platonic representation hypothesis~\citep{huh2024platonic}, both vocabularies are projections of a shared space of dynamical capabilities, and the game is what forces that projection into view.

Taken together, this game-mediated account of meaning and substrate-specific capability connects to older questions about the origin of meaning and the relation between signs, symbols, and their material implementation. Those questions have been developed not only for humans, advanced animals, and societies~\citep{sebeok1965animal, osgood1954psycholinguistics}, but also for minimal systems, biosemiotics, and the emergence of representation in evolutionary contexts~\citep{barbieri2018code, barbieri2008biosemiotics, barbieri2003biology, barbieri1998organic, peirce1967papers, pattee2001physics, pattee1989measurement, pattee1982cell, rosen1985anticipatory}. The Language Game framework extends this tradition in an engineering direction: rather than only asking where meaning comes from, it provides empirical tools for eliciting and testing sense-making capacities in systems where those capacities may not be obvious.

One immediate implication is biomedical. If cells, tissues, organs, and molecular pathways can be treated as nested agents with trainable states, priors, and goals, then communication with those agents becomes a route to top-down control that complements molecular micromanagement~\citep{levin2024multiscale, mathews2023cellular, lagasse2023future}. Such a view is relevant to regenerative medicine, birth defects, aging, cancer, neuroscience, and immunology, where pathway-level learning, memory, forgetting, and bioelectric control already suggest intervention points beyond direct rewiring of molecular hardware~\citep{keresztes2025cancer, veres2024cellular, csermely2020learning, biswas2022learning, biswas2021gene, levin2021bioelectric}. Being able to talk to organs or to molecular pathways inside cells would make explicit a frontier currently approached only indirectly by mind-body medicine, placebo and nocebo research, and studies of language-like effects on physiology~\citep{dossett2020new, black2019mindfulness, taylor2010topdown, shang2001emerging, buric2017molecular, roush1997benson, benson1997nocebo, cannon1957voodoo, evers2018placebo, piedimonte2016words}.

Beyond biomedicine, the same idea points toward bioengineering and hybrid intelligence. Living, designed, and hybrid agential materials---including engineered multicellular systems, synthetic morphologies, biobots, cyborgs, and hybrots---will require interfaces that address their competencies rather than only their parts~\citep{clawson2023endless, kamm2018promise, teague2016synthetic, davies2023synthetic}. The framework may also be useful for artificial and virtual agents, because it lets investigators step outside a system's programmed interface or assumed problem space and probe its abilities in new domains~\citep{zhang2024sorting}. More generally, if both ``thoughts'' and ``thinkers'' can be treated as patterns in excitable systems~\citep{fields2025thoughts}, then language games may become interfaces for many kinds of dynamical system, embodied or otherwise. In this sense, the framework occupies the center of a bowtie architecture: diverse intelligences can appear on either side, and the game-mediated interface can act as cognitive glue~\citep{levin2024memory, levin2023bioelectric}, enabling not only human--system dialogue but also cooperation among non-human intelligences and the engineering of chimeric systems~\citep{nanos2021chimerism}.

These broader connections also clarify the framework's boundaries. A game bounds the vocabulary it can express. A single-agent environment cannot yield words for ``you'', ``other'' or ``they'', since those referents are absent from the underlying game. To reach such relational and social concepts, the framework must be extended to multi-agent and eventually open-ended games; the non-human systems of greatest scientific interest---biological collectives, ecological communities, human organizations---are themselves embedded in games of the relevant richness.

Deploying the framework at biological scale also carries a logistic cost: reinforcement learning demands many evaluations, and at full physical scale direct training is prohibitive; policies also remain memoryless, so dialogue cannot compound across turns. These limits point forward in two directions. Extending beyond ODE models will depend on accurate forward models of more complex organisms, which is itself a prerequisite for reaching the biological substrates of greatest interest. The same cost presses on the learning machinery: scaling dialogue to richer systems will likely demand reinforcement learning methods adapted to expensive environments, and perhaps approaches that step outside the reinforcement learning frame altogether.

These limits notwithstanding, what the Language Game puts forward is, at bottom, a new and practicable paradigm---one in which humans can hold dialogue with arbitrary dynamical systems, and can facilitate communication among a wide range of systems themselves. Language occupies a foundational place in human culture, and once its use is extended to substrates with which we could not previously converse, the reach of such a framework is unlikely to remain confined to engineering or machine learning. More broadly, methods that promote communication among diverse kinds of minds may have positive implications not only for technology, but also for ethics. As rapid technological and biological change expands the space of possible beings, flourishing across the biosphere will depend on tools that reduce mind-blindness and enhance mutual understanding among systems of highly diverse composition and provenance~\citep{levin2025ai, kauffman2008sacred}.

\section*{Acknowledgments}
{\sloppy
This publication was made possible through the support of Grant 62212 from the John Templeton Foundation and grant TWCF-2021-20606 from the Templeton World Charity Foundation, Inc. The opinions expressed in this publication are those of the authors and do not necessarily reflect the views of the John Templeton Foundation. This research was also sponsored by the Army Research Office and was accomplished under Grant Number W911NF-24-1-0041. The views and conclusions contained in this document are those of the authors and should not be interpreted as representing the official policies, either expressed or implied, of the Army Research Office or the U.S. Government. The U.S. Government is authorized to reproduce and distribute reprints for Government purposes notwithstanding any copyright notation herein. The authors acknowledge the Tufts University High Performance Compute Cluster (\url{https://it.tufts.edu/high-performance-computing}) which was utilized for the research reported in this paper.
\par}

\bibliography{ref}
\bibliographystyle{plain}

\newpage
\appendix

\section{PPO Training}

We train PPO policies on reinforcement learning environments spanning Classic Control, MuJoCo~\citep{todorov2012mujoco}, Gymnasium Robotics~\citep{gymnasium_robotics2023}, DeepMind Control Suite~\citep{tassa2018deepmind}, and Atari (ALE)~\citep{bellemare2013arcade}. For each episode, we run~$16$ parallel environments with~$512$ steps per rollout. We use VecNormalize to normalize only the rewards (not observations) to stabilize training. The training budget varies per environment (see Section~\ref{sec:convergence-analysis}); all other hyperparameters are shared and listed in Table~\ref{tab:ppo-hyperparameters}.

\begin{table}[h]
\centering
\begin{tabular*}{\textwidth}{@{\extracolsep{\fill}}lrlr@{}}
\toprule
\textbf{Parameter} & \textbf{Value} & \textbf{Parameter} & \textbf{Value} \\
\midrule
Parallel environments & 16 & Discount factor (~$\gamma$) & 0.99 \\
Rollout steps (~$n_\text{steps}$) & 512 & GAE lambda (~$\lambda$) & 0.95 \\
Batch size & 256 & Clip range & 0.2 \\
Number of epochs & 10 & Reward clipping & 10.0 \\
Learning rate &~$5 \times 10^{-4}$ & Observation clipping & 10.0 \\
Entropy coefficient & 0.0 & Reward normalization & True \\
 & & Observation normalization & False \\
\bottomrule
\end{tabular*}
\caption{PPO training hyperparameters shared across all environments.}
\label{tab:ppo-hyperparameters}
\end{table}

\section{Convergence Analysis and Training Budget Calibration}
\label{sec:convergence-analysis}

Different environments exhibit vastly different learning dynamics: simple tasks such as CartPole converge within a few hundred thousand steps, whereas locomotion tasks like HalfCheetah may require millions. To ensure fair comparison across the GRN--environment matrix, we calibrate a per-environment training budget through the following procedure.

\paragraph{Baseline calibration runs.}
We select the MLP control as the baseline reservoir and train it on each active environment with a generous initial budget---500\,k steps for classic control tasks, 2\,M for medium-complexity environments (Pendulum, MuJoCo, Gymnasium Robotics, DeepMind Control Suite), 3\,M for locomotion tasks (Hopper, HalfCheetah), and 15\,M for Atari RAM games (whose longer episode horizons require more samples). All other hyperparameters remain identical to those in Table~\ref{tab:ppo-hyperparameters}.

\paragraph{Convergence detection.}
For each calibration run, we collect the per-episode reward trace from the 16 parallel monitor logs and sort episodes by cumulative environment steps. We then partition the trace into 50 equal-width bins along the step axis and compute the mean reward per bin. After smoothing with a 5-bin moving average, we measure the normalized improvement between adjacent bins:
\[
  \delta_i = \frac{|\bar{r}_{i+1} - \bar{r}_i|}{r_{\max} - r_{\min}},
\]
where~$r_{\max}$ and~$r_{\min}$ are the global maximum and minimum of the smoothed curve. An environment is declared \emph{converged} at bin~$i$ if~$\delta_j < 0.01$ (less than 1\% of the total reward range per bin) for all~$j \in [i, i + p)$, where the patience window~$p = \lceil 0.15 \times 50 \rceil = 8$ bins. If no such window exists, the environment is classified as either \emph{still rising} (if the mean reward in the last quarter exceeds that of the third quarter by more than 3\% of the range) or \emph{flat/oscillating}.

\paragraph{Budget assignment.}
For converged environments, the training budget is chosen as an empirical safety margin above the detected convergence step, rounded to a convenient boundary and with a minimum of 200\,k steps. In practice this margin is task-dependent: most non-Atari environments use a modest buffer of roughly~$1.2$--~$1.3\times$, while longer-horizon Atari runs use a larger buffer when their learning curves remain noisier near the plateau. For environments still rising at the end of the calibration run, the budget is set conservatively above the calibration budget. The resulting per-environment budgets are listed in Table~\ref{tab:convergence-budgets}.

\begin{table}[h]
\centering
\small
\begin{tabular*}{\textwidth}{@{\extracolsep{\fill}}llrr@{}}
\toprule
\textbf{Category} & \textbf{Environment} & \textbf{Conv.\ step} & \textbf{Budget} \\
\midrule
\multirow{4}{*}{Classic}
  & CartPole-v1               & 502\,k & 650\,k   \\
  & Acrobot-v1                & 107\,k & 200\,k   \\
  & MountainCarContinuous-v0  & 491\,k & 600\,k   \\
  & Pendulum-v1               & 782\,k & 1\,M     \\
\cmidrule(l){2-4}
\multirow{6}{*}{MuJoCo~\citep{todorov2012mujoco}}
  & Reacher-v4                & 461\,k & 600\,k   \\
  & Pusher-v4                 & 702\,k & 900\,k   \\
  & Swimmer-v4                & 1.98\,M& 2.5\,M   \\
  & Hopper-v4                 & 2.98\,M& 3.7\,M   \\
  & HalfCheetah-v4            & 4.94\,M& 6.2\,M   \\
  & HumanoidStandup-v4        & 5\,M   & 6\,M     \\
\cmidrule(l){2-4}
Gymnasium Robotics~\citep{gymnasium_robotics2023}
  & PointMaze\_UMaze-v3       & 421\,k & 550\,k   \\
\cmidrule(l){2-4}
DeepMind Control Suite~\citep{tassa2018deepmind}
  & finger-spin               & 1.62\,M& 2\,M     \\
\cmidrule(l){2-4}
\multirow{4}{*}{Atari (ALE)~\citep{bellemare2013arcade}}
  & BankHeist-ram             & 7.0\,M & 10.5\,M  \\
  & KungFuMaster-ram          & 5\,M   & 8\,M     \\
  & CrazyClimber-ram          & 7\,M   & 10\,M    \\
  & Kangaroo-ram              & 5\,M   & 8\,M     \\
\bottomrule
\end{tabular*}
\caption{Per-environment training budgets calibrated from MLP convergence analysis. Budgets are empirical safety margins above the detected convergence step, rounded to convenient step counts, with a minimum of 200\,k.}
\label{tab:convergence-budgets}
\end{table}

\paragraph{Observations.}
All 16 environments in the calibration table converge reliably under MLP training. Seven converge within 1\,M steps (CartPole, Acrobot, MountainCarContinuous, Pendulum, Reacher, Pusher, PointMaze), and the remaining nine converge between 1.6\,M and 7\,M steps. HumanoidStandup plateaus around 5\,M steps (reward oscillates between 148\,k--160\,k thereafter). All four Atari RAM environments converge reliably: KungFuMaster and Kangaroo at 5\,M, BankHeist and CrazyClimber at 7\,M.

\section{Gene Regulatory Network Models}
\label{sec:grn-models}

We use 14 GRN models sourced from OdeBase (\url{https://odebase.org}), spanning six biological categories. Together with two non-biological controls (\texttt{identity} and \texttt{mlp}) and the Lorenz system baseline, these form the 17 reservoirs in our evaluation matrix. Table~\ref{tab:grn-models} lists all models with their state-space dimensions and biological context.

\begin{table}[h]
\centering
\small
\begin{tabular}{llcll}
\hline
\textbf{Category} & \textbf{Model} & \textbf{Dim} & \textbf{Biological Process} & \textbf{OdeBase} \\[0.5ex]
\hline
\multirow{3}{*}{Baseline / Control}
  & LorenzSystem              & 3  & Chaotic attractor (non-biological) & --- \\
  & \texttt{identity}         &~$d$ & Identity mapping (control)        & --- \\
  & \texttt{mlp}              &~$d$ & 2-layer MLP (control)             & --- \\[0.3ex]
\hline
\multirow{4}{*}{Circadian Clock}
  & Tyson1999CircleLock                     & 3  & Circadian limit cycle~\citep{tyson1999simple}                  & 1351 \\
  & Weimann2004CircadianOscillator          & 7  & Mammalian circadian oscillator~\citep{beckerweimann2004circadian} & 1432 \\
  & Almeida2019CircadianClock               & 8  & Neurospora circadian clock~\citep{almeida2020circadian}         & 1850 \\
  & Leloup1999CircadianClock                & 10 & Drosophila PER/TIM clock~\citep{leloup1999chaos}                & 1338 \\[0.3ex]
\hline
\multirow{3}{*}{Cell Cycle}
  & Tyson1991CellCycle2Var                  & 4  & Fission yeast cell cycle~\citep{tyson1991modeling}             & 1328 \\
  & Gardner1998CellCycleGoldbeter           & 5  & Goldbeter minimal cascade~\citep{gardner1998cellcycle}         & 1329 \\
  & Gerard2010CellCycle                     & 8  & Mammalian Cdk oscillator~\citep{gerard2011skeleton}            & 1928 \\[0.3ex]
\hline
\multirow{2}{*}{Cell Fate}
  & Chickarmane2008NanogGata6               & 6  & NANOG/GATA6 lineage switch~\citep{chickarmane2008computational} & 1456 \\
  & Chickarmane2006StemCellSwitch            & 12 & Stem cell pluripotency circuit~\citep{chickarmane2006stemcell}  & 1450 \\[0.3ex]
\hline
\multirow{2}{*}{Signal Transduction}
  & Markevich2004MAPK                       & 5  & MAPK double phosphorylation~\citep{markevich2004signaling}     & 1343 \\
  & Kholodenko2000MAPKCascade               & 8  & MAPK 3-tier cascade~\citep{kholodenko2000mapk}                 & 1331 \\[0.3ex]
\hline
p53 / DNA Damage
  & Zatorsky2006P53Model4                   & 3  & p53-Mdm2 oscillator~\citep{gevazatorsky2006p53}                & 1422 \\[0.3ex]
\hline
\multirow{2}{*}{Synthetic Circuit}
  & Gardner2000ToggleSwitch                 & 3  & E.\ coli genetic toggle switch~\citep{gardner2000toggle}       & 1646 \\
  & Liebal2012TranscriptionInhibition       & 4  & B.\ subtilis transcription~\citep{liebal2012proteolysis}       & 1616 \\
\hline
\end{tabular}
\caption{Reservoir models used in the evaluation matrix. \textbf{Dim}: number of ODE state variables (reservoir dimension). The \texttt{identity} and \texttt{mlp} controls use a reservoir dimension~$d = 256$. \textbf{OdeBase}: identifier on \url{https://odebase.org/detail/<id>}. All GRN models are implemented as differentiable PyTorch modules with frozen parameters; only the linear encoder and decoder layers are trained. The implementation also includes batch normalization layers around the reservoir and, for continuous-action environments, a learnable log-standard-deviation parameter; these auxiliary parameters are trained alongside~$E$ and~$D$.}
\label{tab:grn-models}
\end{table}

The GRN dimensions range from 3 (Lorenz, Tyson1999, Zatorsky2006, Gardner2000) to 12 (Chickarmane2006), providing a natural experiment in how reservoir dimensionality and dynamical structure jointly affect policy learning. Circadian clock models (3--10 dimensions) exhibit limit-cycle dynamics, cell cycle models (4--8 dimensions) feature bistable switches and oscillations, and signal transduction cascades (5--8 dimensions) implement ultrasensitive responses---all qualitatively different dynamical repertoires that the linear readout must learn to exploit.

\begin{table}[h]
\centering
\small
\begin{tabular*}{\textwidth}{@{\extracolsep{\fill}}llp{0.55\textwidth}@{}}
\toprule
\textbf{Category} & \textbf{Property} & \textbf{GRN Models} \\
\midrule
\multirow{5}{*}{\shortstack[l]{Dynamics /\\Constraint}}
  & Oscillatory      & Tyson99, Weimann04, Almeida19, Leloup99, Tyson91, Gardner98, Gerard10, Zatorsky06, Kholodenko00 \\
  & Bistable         & Gardner00, Chickarmane06, Chickarmane08 \\
  & Neg.\ Feedback   & Tyson99, Weimann04, Almeida19, Leloup99, Zatorsky06, Kholodenko00 \\
  & Ultrasensitivity & Markevich04, Gardner98, Gerard10, Gardner00 \\
  & Non-oscillatory  & Chickarmane06, Chickarmane08, Gardner00, Liebal12, Markevich04 \\
\midrule
\multirow{5}{*}{\shortstack[l]{Biological\\Context}}
  & Circadian        & Tyson99, Weimann04, Almeida19, Leloup99 \\
  & Cell Cycle       & Tyson91, Gardner98, Gerard10 \\
  & Cell Fate        & Chickarmane06, Chickarmane08 \\
  & Signal Transd.   & Markevich04, Kholodenko00, Almeida19 \\
  & Transcriptional  & Almeida19, Liebal12, Chickarmane06, Chickarmane08 \\
\midrule
\multirow{3}{*}{\shortstack[l]{Mechanism /\\Constraint}}
  & Phosphorylation  & Markevich04, Kholodenko00, Leloup99, Gerard10, Gardner98 \\
  & Complex Form.    & Almeida19, Leloup99, Chickarmane06 \\
  & Conservation     & Tyson91, Markevich04, Kholodenko00 \\
\bottomrule
\end{tabular*}
\caption{GRN property taxonomy. The binary GRN properties are assigned from OdeBase model metadata and source links, source model names, original papers, ODE/reaction-network structure, and relevant systems-biology knowledge. They combine biological context, molecular mechanism, dynamical behavior, and model-level constraints. A GRN can carry multiple properties across categories. Short names refer to models in Table~\ref{tab:grn-models}.}
\label{tab:grn-properties}
\end{table}

\paragraph{\texorpdfstring{Property meanings and selection rationale.}{Property meanings and selection rationale.}}
Table~\ref{tab:grn-property-rationale} summarizes why each GRN property was selected and how it was assigned. The evidence comes from OdeBase/source model names, original paper titles, model categories, species and reaction names, and explicit constraints in the implemented ODEs.

\begingroup
\footnotesize
\begin{longtable}{p{0.18\textwidth}p{0.30\textwidth}p{0.42\textwidth}}
\caption{GRN property meanings, selection rationale, and evidence.}
\label{tab:grn-property-rationale}\\
\toprule
\textbf{Property} & \textbf{Meaning} & \textbf{Selection rationale and evidence} \\
\midrule
\endfirsthead
\caption[]{GRN property meanings, selection rationale, and evidence (continued).}\\
\toprule
\textbf{Property} & \textbf{Meaning} & \textbf{Selection rationale and evidence} \\
\midrule
\endhead
\bottomrule
\endfoot
Oscillatory & Clock, pulse, cell-cycle, or signaling models with recurrent dynamics. & Chosen to capture timing and limit-cycle-like reservoirs. Assigned when the OdeBase/source name, paper title, or model category identifies a clock, cell cycle, p53 oscillation, or MAPK oscillation. \\
Bistable & Models with two stable expression or activation regimes. & Chosen to capture switch-like memory and state commitment. Assigned from model names such as genetic toggle switch, stem-cell switch, and NANOG/GATA6 lineage switch. \\
Neg.\ Feedback & Models containing an explicit negative feedback loop. & Chosen because feedback is a core mechanism for regulation and oscillation. Assigned when model names, species interactions, or ODE terms show repression or inhibitory feedback, such as p53--Mdm2 or MAPK-PP feedback. \\
Ultrasensitivity & Models with steep, switch-like input--output responses. & Chosen to capture high-gain nonlinear motifs. Assigned from mechanisms such as multisite phosphorylation, Goldbeter--Koshland switches, Cdk cascades, and synthetic toggle-switch kinetics. \\
Non-oscillatory & Models treated as switches, steady transcription modules, or signaling modules rather than clocks. & Chosen as a contrast to oscillatory reservoirs. Assigned to stem-cell switches, the genetic toggle, transcription inhibition, and the Markevich MAPK double-phosphorylation switch. \\
Circadian & Molecular circadian clock models. & Chosen to group reservoirs evolved for biological timing. Assigned directly from OdeBase/source names and model categories containing circadian clock or circadian oscillator. \\
Cell Cycle & Models of cyclin, Cdc2/Cdk, or related cell-cycle control. & Chosen because these networks combine oscillation, checkpoints, and kinase regulation. Assigned from source categories and model names containing cell cycle. \\
Cell Fate & Models of lineage commitment, pluripotency, or stem-cell fate decisions. & Chosen to represent attractor-like biological decision circuits. Assigned from stem-cell switch and NANOG/GATA6 lineage model names and species. \\
Signal Transd. & Models of signaling pathways or signaling-responsive clock modules. & Chosen to capture reservoirs that transform upstream inputs into downstream states. Assigned to MAPK models and to the Almeida circadian model because its source title emphasizes signaling inputs. \\
Transcriptional & Models that explicitly include transcription, mRNA production, promoters, transcription factors, or gene-expression inhibition. & Chosen because transcriptional regulation is central to GRN computation. Assigned from model titles and species names such as mRNA, gene, promoter, OCT4, SOX2, NANOG, sigB, and lacZ. \\
Phosphorylation & Models centered on phosphorylation, dephosphorylation, or kinase cascades. & Chosen to capture covalent-modification signal processing. Assigned from MAPK phosphorylation models, PER/TIM phosphorylation in the clock, and cyclin/Cdk cell-cycle kinase modules. \\
Complex Form. & Models that explicitly include protein complexes or dimer states. & Chosen because complex formation couples molecular species into coordinated state variables. Assigned when species or reactions include complexes such as PER/CRY, PER/TIM, or OCT4--SOX2. \\
Conservation & Models with conserved totals or stoichiometric constraints. & Chosen to capture model-level constraints on reachable states. Assigned when the implementation enforces relations such as~$v=u+z$, total MAPK conservation, or conserved total kinase in cascade tiers. \\
\end{longtable}
\endgroup

\section{\texorpdfstring{Per-cell Contingency Tests for (Property, Environment) Pairs}{Per-cell Contingency Tests for (Property, Environment) Pairs}}
\label{sec:contingency}

The universal sign test in Section~\ref{sec:biological-inductive-biases} collapses the entire matrix of environment-level comparisons into a single binomial observation per GRN property, trading localization for power.  The complementary question---\emph{which} environments does each property help or hurt on?---requires a per-cell test that preserves the~$(\text{property}, \text{environment})$ grid.  We address it with a~$2 \times 2$ contingency-table analysis.

\paragraph{Procedure.}
For each environment~$e$ we partition the 14 biological GRNs into a ``good'' half (top seven by seed-averaged final reward) and a ``bad'' half (bottom seven); the median split is exact because~$14 = 2 \times 7$.  For each GRN property~$b$, the same 14 GRNs are independently partitioned into those carrying~$b$ and those without, yielding a~$2 \times 2$ contingency table
\[
\begin{array}{l|cc}
                  & \text{good} & \text{bad} \\\hline
\text{with } b    & n_{11}      & n_{10} \\
\text{without } b & n_{01}      & n_{00}
\end{array}
\qquad (n_{11} + n_{01} = n_{10} + n_{00} = 7,\;\; n_{11} + n_{10} = |G_b|,\;\; n_{01} + n_{00} = 14 - |G_b|).
\]
We run a two-sided Fisher exact test on each of the~$13 \times 16 = 208$ tables.  For visualisation, we report a Haldane--Anscombe-corrected log odds ratio~$\log\bigl[(n_{11} + 0.5)(n_{00} + 0.5) / ((n_{10} + 0.5)(n_{01} + 0.5))\bigr]$ that remains finite when any cell is empty.  Benjamini--Hochberg FDR correction is applied jointly across all 208 cells.

\paragraph{Statistical ceiling.}
The discrete Fisher distribution imposes a hard lower bound on attainable~$p$-values that depends on the property marginal.  With~$|G_b| = k$ and a balanced~$7/7$ split of 14 GRNs, the two-sided~$p$-value from the most extreme table is
\[
    p_\text{min}(k) \;=\; 2 \cdot \frac{\binom{k}{k}\binom{14 - k}{7 - k}}{\binom{14}{7}},
\]
which evaluates to~$0.192$,~$0.070$,~$0.021$ and~$0.005$ for~$k \in \{3, 4, 5, 6\}$ respectively (and symmetrically for~$k \in \{11, 10, 9, 8\}$).  Eight of the thirteen properties have~$k \in \{3, 4\}$, so for these rows no individual cell can cross the conventional~$\alpha = 0.05$ threshold even in the most extreme configuration.  The analysis is therefore deliberately conservative: it uses the finest available spatial resolution at the cost of nominal cell-level significance.

\begin{figure}[tb]
    \centering
    \includegraphics[width=\textwidth]{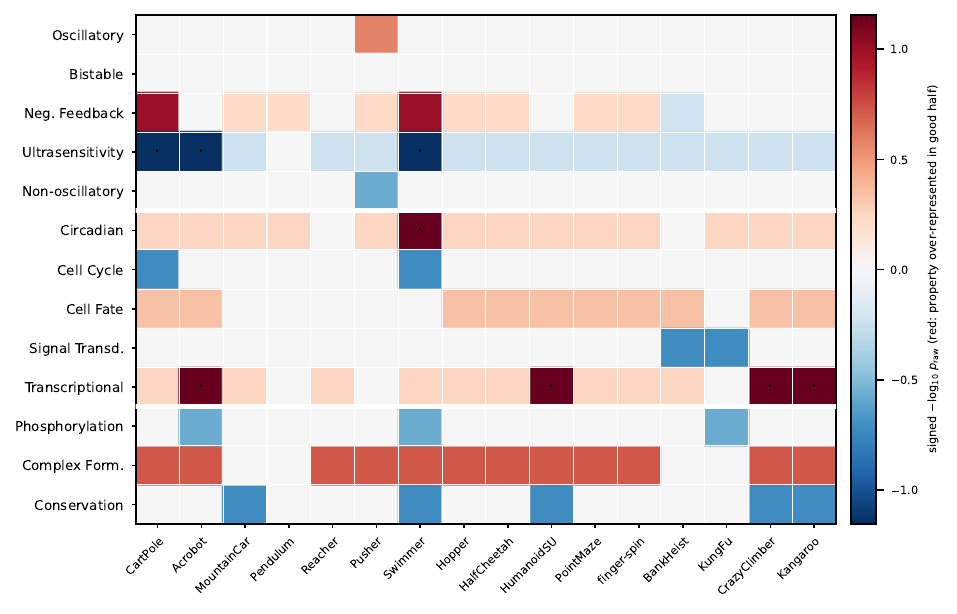}
    \caption{Per-cell~$2 \times 2$ contingency-table heatmap.  Each cell corresponds to one (GRN property, environment) pair; color encodes the signed~$-\log_{10}$ of the raw two-sided Fisher-exact~$p$-value, with red indicating that the property is over-represented in the top-seven (good) half of GRNs on that environment and blue that it is over-represented in the bottom-seven (bad) half.  Small dots mark cells with nominal~$p < 0.10$; stars would mark cells surviving Benjamini--Hochberg FDR correction at~$q < 0.05$ across all 208 tests (none do, for the discreteness reasons discussed in the text).  Category separators match those used in Figure~\ref{fig:bio-priors}.}
    \label{fig:contingency-heatmap}
\end{figure}

\paragraph{Results.}
Figure~\ref{fig:contingency-heatmap} displays the signed~$-\log_{10} p_\text{raw}$ for every (property, environment) pair, with stars reserved for cells that survive FDR correction and a small dot marking cells that are nominally interesting (~$p_\text{raw} < 0.10$).  No cell crosses~$q < 0.05$, consistent with the saturation bound above: the eight cells with smallest~$p$-values all sit at~$p = 0.070$, the ceiling for~$k = 4$ properties.  Despite this, the nominally interesting cells form coherent rows that reproduce the sign-test findings at a finer spatial resolution.  \emph{Ultrasensitivity} (~$k = 4$) collapses into the bad half on CartPole, Acrobot and Swimmer; \emph{Transcriptional} (~$k = 4$) occupies the good half on Acrobot, HumanoidStandup, CrazyClimber and Kangaroo; \emph{Circadian} (~$k = 4$) is over-represented in the good half on Swimmer.  Rows with large~$k$ (Oscillatory~$k = 9$, Neg.\ Feedback~$k = 6$) show weaker contrasts because their partitions are close to balanced.

\paragraph{Interpretation.}
The contingency heatmap should be read as a localization map rather than a confirmatory test.  Because the discrete null distribution places the attainable floor for most rows above the~$\alpha = 0.05$ mark, absence of a star does not imply absence of an effect; conversely, the spatial clustering of nominally interesting cells within rows---and the agreement of their direction with the universal sign test---is itself evidence that the properties carry environment-specific information beyond the single aggregate sign summarized in Section~\ref{sec:biological-inductive-biases}.  A decisive per-cell analysis would require substantially more than 14 biological GRNs per partition.

\section{Full Training Reward Curves}
\label{sec:full-reward-curves}

\begin{figure}[p!]
    \centering
    \includegraphics[width=0.85\textwidth,height=0.88\textheight,keepaspectratio]{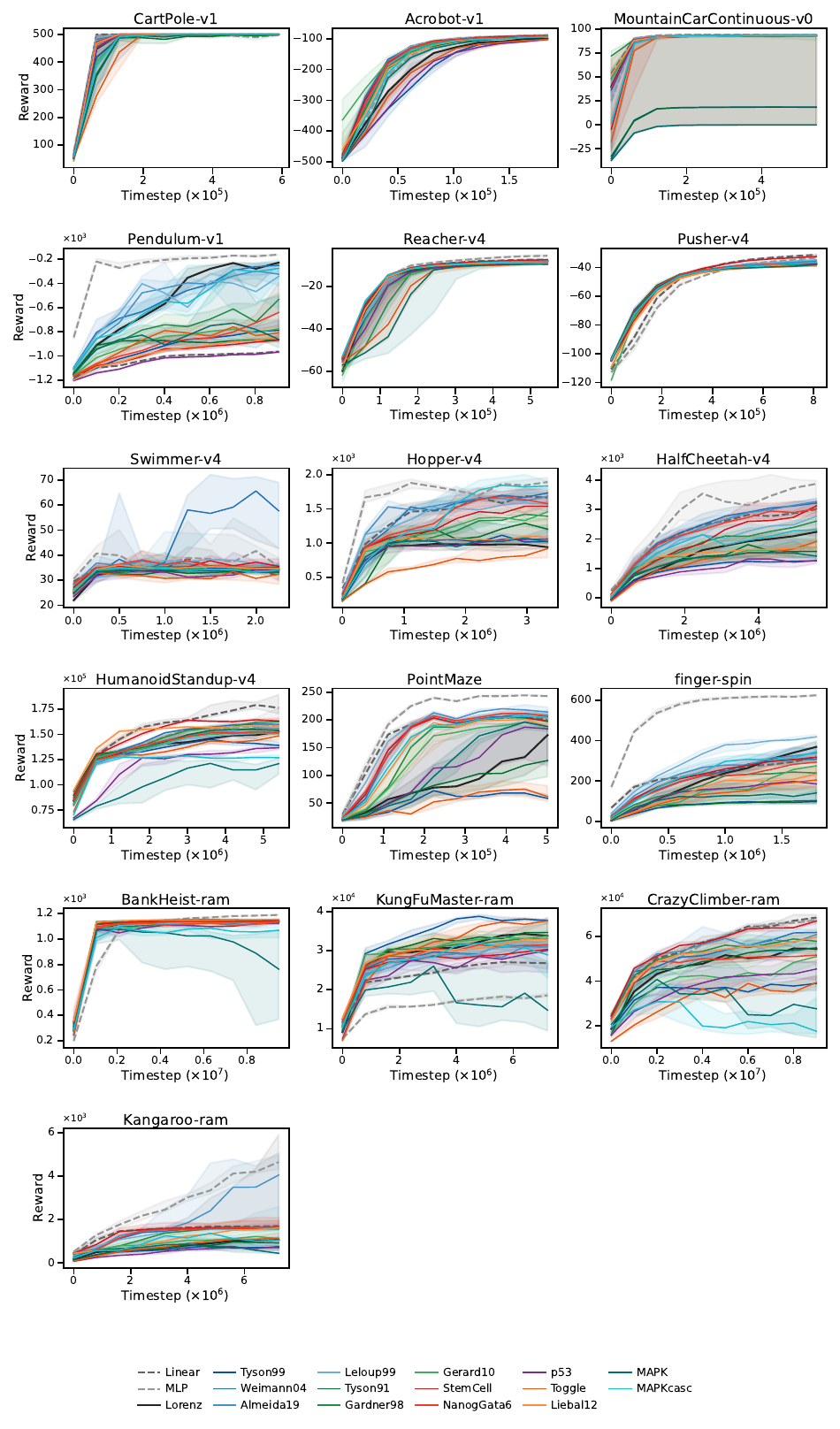}
    \caption{Training reward curves for all 16 RL environments with 17 reservoir architectures. Dashed lines indicate control baselines (\texttt{identity}, \texttt{mlp}); solid lines indicate GRN-based and Lorenz reservoirs. Colors are grouped by biological category (see legend). Shaded regions show the 70th--90th percentile range across seeds.}
    \label{fig:rewards-full}
\end{figure}

Figure~\ref{fig:rewards-full} presents the complete training reward curves for all 16 reinforcement learning environments across all 17 reservoirs. This extends the representative subset shown in Figure~\ref{fig:rewards} in the main text.

\section{Reinforcement Learning Environments}
\label{sec:rl-environments}

Table~\ref{tab:env-catalog} provides a comprehensive catalog of the 16 reinforcement learning environments used in our evaluation matrix. For each environment we list the observation and action space dimensions, reward structure, episode horizon, and the criterion by which the task is considered solved. Figure snapshots show the final frame of a trained MLP-baseline agent executing its learned policy, illustrating the visual diversity of the benchmark suite.

We report the user-facing environment names used throughout the paper. At runtime, these names are resolved to the installed Gymnasium-compatible IDs: \texttt{Pusher-v4} is run as \texttt{Pusher-v5} because the older MuJoCo-v4 environment depends on older MuJoCo bindings; \texttt{PointMaze} resolves to \texttt{PointMaze\_UMaze-v3}; and the Atari names with the \texttt{-ram} suffix resolve to ALE-v5 games with \texttt{obs\_type="ram"}. The aliasing affects package compatibility only, not the semantic task labels used in the language-game pipeline.

\begin{small}
\setlength{\LTpre}{4pt}
\setlength{\LTpost}{4pt}
\begin{longtable}{p{0.19\textwidth} p{0.77\textwidth}}
\caption{Reinforcement learning environment catalog. Each entry shows a rendered snapshot of the trained MLP-baseline agent together with the environment's meta-information and solved criterion.}
\label{tab:env-catalog} \\
\toprule
\textbf{Snapshot} & \textbf{Description} \\
\midrule
\endfirsthead
\multicolumn{2}{l}{\small\textit{(Table~\ref{tab:env-catalog} continued)}} \\
\toprule
\textbf{Snapshot} & \textbf{Description} \\
\midrule
\endhead
\midrule
\multicolumn{2}{r}{\small\textit{Continued on next page}} \\
\endfoot
\bottomrule
\endlastfoot


\raisebox{-0.9\height}{\includegraphics[width=\linewidth]{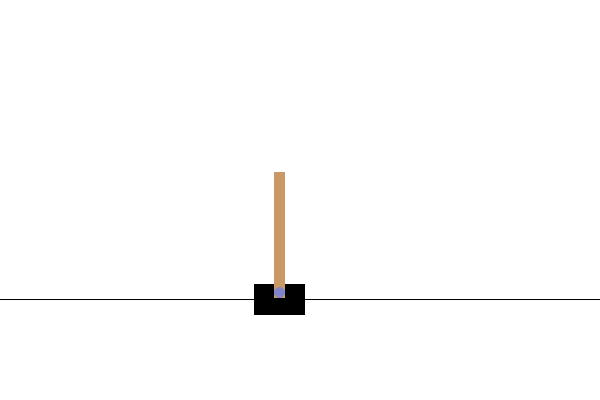}}
&
\textbf{CartPole-v1} \hfill \textit{Classic Control} \par\vspace{2pt}
\textbf{Task:} Balance a pole hinged on a moving cart by pushing the cart left or right. The agent must keep the pole upright for as long as possible without letting it fall or the cart leave the track. \par
\textbf{Obs:}~$\mathbb{R}^{4}$ (cart position, velocity, pole angle, angular velocity). \par
\textbf{Action:} Discrete,~$|\mathcal{A}|=2$ (push left / right). \par
\textbf{Reward:}~$+1$ per step the pole remains upright (dense). \par
\textbf{Horizon:} 500 steps. \par
\textbf{Solved:} Average return~$\geq 475$ over 100 episodes, i.e.\ the pole stays balanced for nearly the entire episode.
\\ \midrule

\raisebox{-0.9\height}{\includegraphics[width=\linewidth]{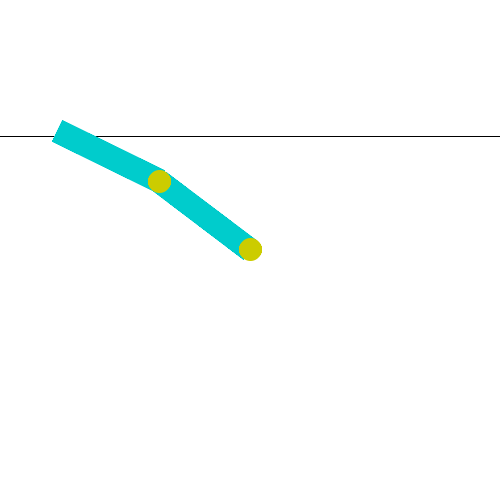}}
&
\textbf{Acrobot-v1} \hfill \textit{Classic Control} \par\vspace{2pt}
\textbf{Task:} Swing the tip of a two-link pendulum above a target height by applying torque only at the middle joint. The system starts hanging downward; the agent must build momentum through under-actuated swings. \par
\textbf{Obs:}~$\mathbb{R}^{6}$ (~$\cos\theta_1, \sin\theta_1, \cos\theta_2, \sin\theta_2, \dot\theta_1, \dot\theta_2$). \par
\textbf{Action:} Discrete,~$|\mathcal{A}|=3$ (torque~$\in\{-1,0,+1\}$). \par
\textbf{Reward:}~$-1$ per step;~$0$ upon reaching the target height (dense). \par
\textbf{Horizon:} 500 steps. \par
\textbf{Solved:} Average return~$\geq -100$ over 100 episodes, meaning the tip reaches the target height within~$\sim$100 steps on average.
\\ \midrule

\raisebox{-0.9\height}{\includegraphics[width=\linewidth]{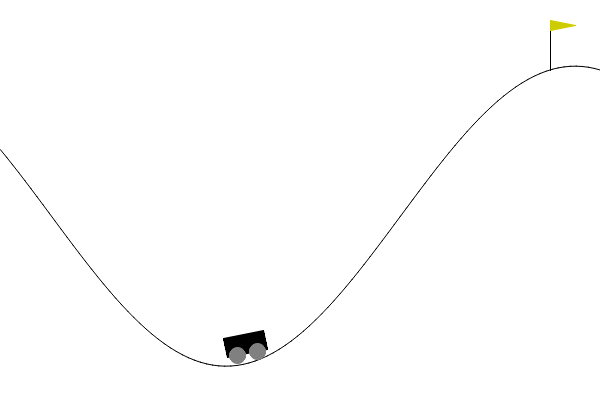}}
&
\textbf{MountainCarContinuous-v0} \hfill \textit{Classic Control} \par\vspace{2pt}
\textbf{Task:} Drive an under-powered car up a steep hill by building momentum through back-and-forth oscillations. The engine is too weak to climb directly; the agent must exploit the valley's slope. \par
\textbf{Obs:}~$\mathbb{R}^{2}$ (position, velocity). \par
\textbf{Action:} Continuous,~$\mathbb{R}^{1}$ (force~$\in[-1,1]$). \par
\textbf{Reward:}~$+100$ for reaching the hilltop,~$-0.1\,a^2$ per step (sparse + shaping). \par
\textbf{Horizon:} 999 steps. \par
\textbf{Solved:} Average return~$\geq 90$ over 100 episodes, meaning the car consistently reaches the goal with minimal energy expenditure.
\\ \midrule

\raisebox{-0.9\height}{\includegraphics[width=\linewidth]{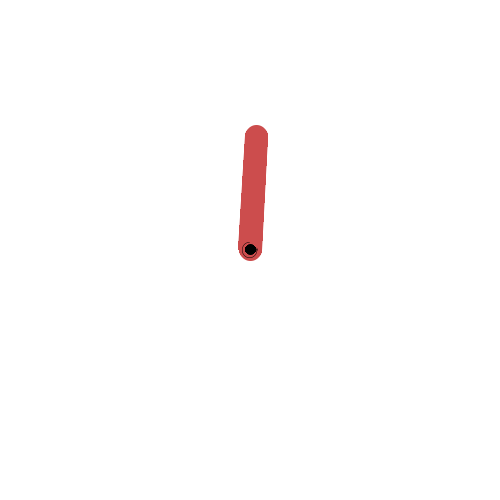}}
&
\textbf{Pendulum-v1} \hfill \textit{Classic Control} \par\vspace{2pt}
\textbf{Task:} Swing up a frictionless pendulum from a random initial angle and stabilize it in the inverted (upright) position using bounded torque. \par
\textbf{Obs:}~$\mathbb{R}^{3}$ (~$\cos\theta, \sin\theta, \dot\theta$). \par
\textbf{Action:} Continuous,~$\mathbb{R}^{1}$ (torque~$\in[-2,2]$). \par
\textbf{Reward:}~$-(\theta^2 + 0.1\,\dot\theta^2 + 0.001\,a^2)$; range~$[-16.27,\;0]$ (dense). \par
\textbf{Horizon:} 200 steps. \par
\textbf{Solved:} No official threshold; a return near~$0$ indicates the pendulum is held upright with minimal oscillation and control effort throughout the episode.
\\ \midrule


\raisebox{-0.9\height}{\includegraphics[width=\linewidth]{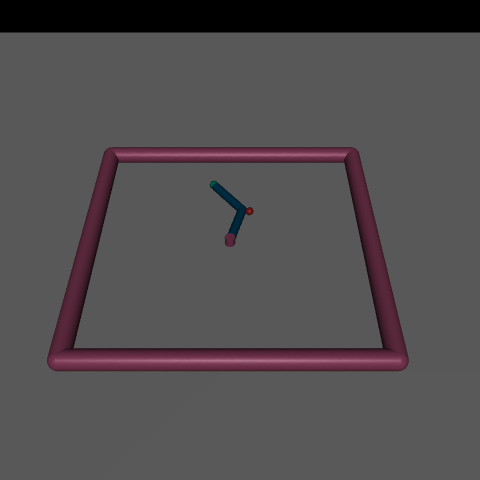}}
&
\textbf{Reacher-v4} \hfill \textit{MuJoCo} \par\vspace{2pt}
\textbf{Task:} Control a planar two-link robotic arm to move its fingertip to a randomly placed target. The agent must learn accurate inverse kinematics through trial and error. \par
\textbf{Obs:}~$\mathbb{R}^{11}$ (joint angles, velocities, fingertip-to-target vector). \par
\textbf{Action:} Continuous,~$\mathbb{R}^{2}$ (joint torques). \par
\textbf{Reward:}~$-\|$fingertip~$-$ target~$\|$~$-\|a\|^2$ (dense). \par
\textbf{Horizon:} 50 steps. \par
\textbf{Solved:} Average return~$\geq -3.75$ (community benchmark), meaning the fingertip is consistently close to the target with low control effort.
\\ \midrule

\raisebox{-0.9\height}{\includegraphics[width=\linewidth]{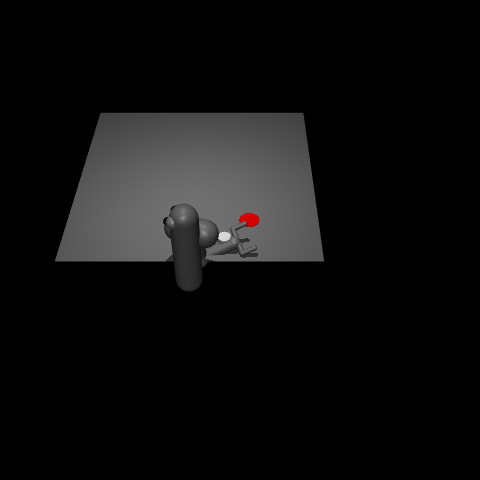}}
&
\textbf{Pusher-v4} \hfill \textit{MuJoCo} \par\vspace{2pt}
\textbf{Task:} Use a 7-DoF robotic arm to push an object on a table toward a goal position. The agent must coordinate reaching the object and then displacing it accurately. \par
\textbf{Obs:}~$\mathbb{R}^{23}$ (arm joints, object and goal positions). \par
\textbf{Action:} Continuous,~$\mathbb{R}^{7}$ (joint torques of 7-DoF arm). \par
\textbf{Reward:}~$-\|$object~$-$ goal~$\| - \|$fingertip~$-$ object~$\| - \|a\|^2$ (dense). \par
\textbf{Horizon:} 100 steps. \par
\textbf{Solved:} No official threshold; lower distance penalties indicate the object is pushed closer to the goal.
\\ \midrule

\raisebox{-0.9\height}{\includegraphics[width=\linewidth]{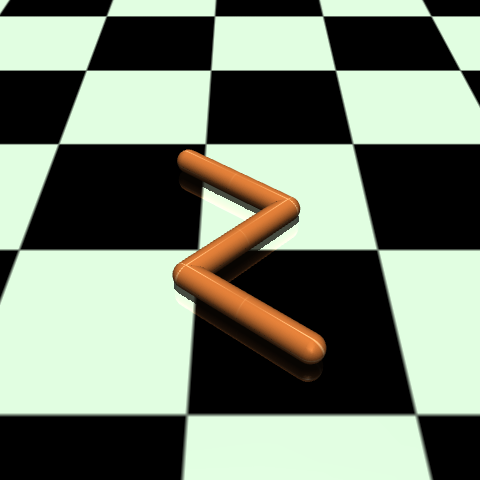}}
&
\textbf{Swimmer-v4} \hfill \textit{MuJoCo} \par\vspace{2pt}
\textbf{Task:} Propel a three-link planar swimmer forward through a viscous fluid by coordinating sinusoidal joint rotations, mimicking biological undulatory locomotion. \par
\textbf{Obs:}~$\mathbb{R}^{8}$ (joint angles, velocities). \par
\textbf{Action:} Continuous,~$\mathbb{R}^{2}$ (rotor torques). \par
\textbf{Reward:} Forward velocity~$- 10^{-4}\|a\|^2$ (dense). \par
\textbf{Horizon:} 1000 steps. \par
\textbf{Solved:} Average return~$\geq 360$ (community benchmark), corresponding to sustained forward swimming at high velocity.
\\ \midrule

\raisebox{-0.9\height}{\includegraphics[width=\linewidth]{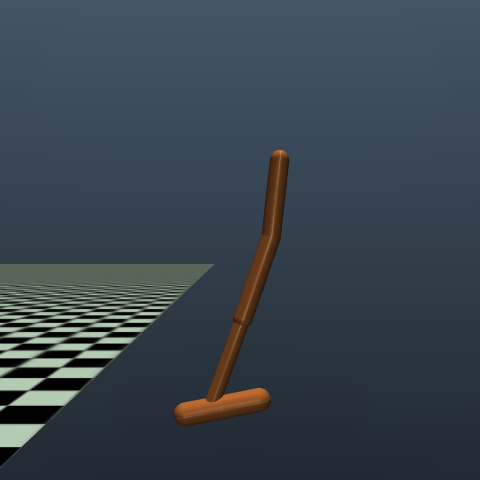}}
&
\textbf{Hopper-v4} \hfill \textit{MuJoCo} \par\vspace{2pt}
\textbf{Task:} Make a one-legged planar robot hop forward as fast as possible without falling over. The agent must coordinate hip, knee, and ankle torques to produce a stable hopping gait. \par
\textbf{Obs:}~$\mathbb{R}^{11}$ (torso height, joint angles, velocities). \par
\textbf{Action:} Continuous,~$\mathbb{R}^{3}$ (hip, knee, ankle torques). \par
\textbf{Reward:} Forward velocity~$+ 1.0$ (alive bonus)~$- 10^{-3}\|a\|^2$ (dense). \par
\textbf{Horizon:} 1000 steps. Terminates early on fall. \par
\textbf{Solved:} Average return~$\geq 3800$ (community benchmark), indicating fast, stable hopping over the full episode.
\\ \midrule

\raisebox{-0.9\height}{\includegraphics[width=\linewidth]{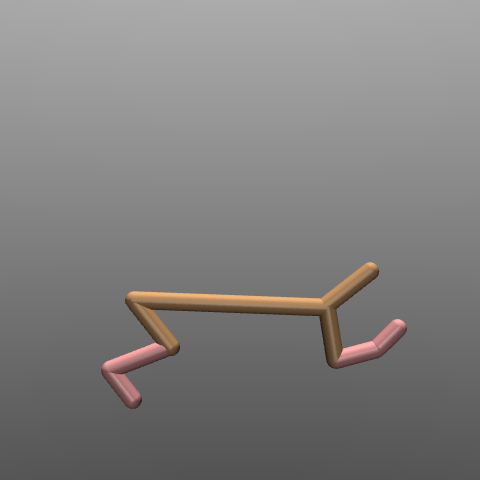}}
&
\textbf{HalfCheetah-v4} \hfill \textit{MuJoCo} \par\vspace{2pt}
\textbf{Task:} Control a planar bipedal robot resembling a running cheetah to sprint forward as fast as possible by coordinating six joint torques. \par
\textbf{Obs:}~$\mathbb{R}^{17}$ (root~$z$-coordinate, joint angles, velocities). \par
\textbf{Action:} Continuous,~$\mathbb{R}^{6}$ (joint torques). \par
\textbf{Reward:} Forward velocity~$- 0.1\,\|a\|^2$ (dense). \par
\textbf{Horizon:} 1000 steps. \par
\textbf{Solved:} Average return~$\geq 4800$ (community benchmark), corresponding to high-speed forward locomotion with efficient actuation.
\\ \midrule

\raisebox{-0.9\height}{\includegraphics[width=\linewidth]{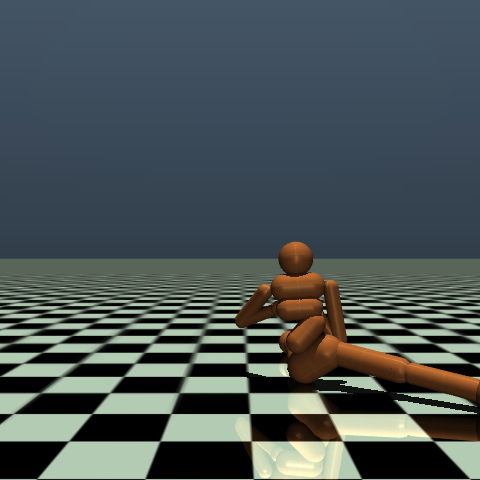}}
&
\textbf{HumanoidStandup-v4} \hfill \textit{MuJoCo} \par\vspace{2pt}
\textbf{Task:} Make a 3D humanoid robot, starting from a prone position on the ground, stand up as tall as possible. This is one of the highest-dimensional control tasks in the MuJoCo suite (17 actuators, 376-dimensional observation). \par
\textbf{Obs:}~$\mathbb{R}^{376}$ (joint positions, velocities, contact forces). \par
\textbf{Action:} Continuous,~$\mathbb{R}^{17}$ (joint torques). \par
\textbf{Reward:} Torso height bonus~$- $ control cost~$- $ contact cost (dense). \par
\textbf{Horizon:} 1000 steps. \par
\textbf{Solved:} No official threshold; good policies exceed~$10^5$ cumulative return, indicating the humanoid reaches and maintains a standing posture.
\\ \midrule


\raisebox{-0.9\height}{\includegraphics[width=\linewidth]{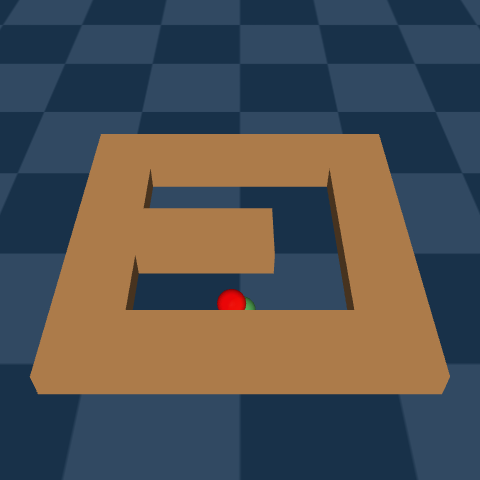}}
&
\textbf{PointMaze\_UMaze-v3} \hfill \textit{Gymnasium Robotics} \par\vspace{2pt}
\textbf{Task:} Navigate a point-mass agent through a U-shaped maze to reach a goal position. The agent must plan a path around the walls rather than moving in a straight line toward the goal. \par
\textbf{Obs:}~$\mathbb{R}^{8}$ (flattened Dict: agent position, velocity, goal position). \par
\textbf{Action:} Continuous,~$\mathbb{R}^{2}$ (velocity commands). \par
\textbf{Reward:} Sparse:~$-1$ per step,~$0$ when within tolerance of goal. \par
\textbf{Horizon:} 300 steps. Terminates early upon reaching the goal. \par
\textbf{Solved:} The agent consistently reaches the goal within the episode; shorter paths (higher returns) indicate better navigation.
\\ \midrule


\raisebox{-0.9\height}{\includegraphics[width=\linewidth]{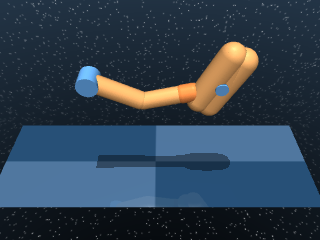}}
&
\textbf{finger-spin} \hfill \textit{DeepMind Control Suite} \par\vspace{2pt}
\textbf{Task:} Use a planar finger with two actuated joints to spin a free-rotating body as fast as possible. The agent must learn to flick the body and maintain continuous rotational momentum. \par
\textbf{Obs:}~$\mathbb{R}^{9}$ (flattened Dict: finger joint positions, velocities, spinner state). \par
\textbf{Action:} Continuous,~$\mathbb{R}^{2}$ (finger joint torques). \par
\textbf{Reward:} Proportional to angular velocity of the spinner;~$\in[0,1]$ per step (dense). \par
\textbf{Horizon:} 1000 steps. \par
\textbf{Solved:} Cumulative return~$\approx 1000$, meaning the spinner is kept at maximum angular velocity throughout the episode.
\\ \midrule


\raisebox{-0.9\height}{\includegraphics[width=\linewidth]{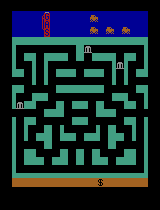}}
&
\textbf{BankHeist-ram} \hfill \textit{Atari (ALE)} \par\vspace{2pt}
\textbf{Task:} Rob banks across a city maze while evading police cars. The player drives through maze corridors, entering banks to collect loot and planting dynamite to destroy pursuing police. Fuel depletes continuously, requiring strategic route planning. New levels add more police and alter maze layouts. \par
\textbf{Obs:}~$\mathbb{R}^{128}$ (console RAM, \texttt{uint8}; semantically encodes player position, police positions, bank states, fuel level, score). \par
\textbf{Action:} Discrete,~$|\mathcal{A}|=18$ (8 directions~$\times$ fire/no-fire + NOOP + fire-only). \par
\textbf{Reward:}~$+10$--~$100$ per bank robbed (level-dependent),~$+20$ per police car destroyed (dense). \par
\textbf{Horizon:} 108{,}000 frames. Terminates when all lives are lost. \par
\textbf{Solved:} No standard threshold; higher scores reflect efficient route planning and evasion.
\\ \midrule

\raisebox{-0.9\height}{\includegraphics[width=\linewidth]{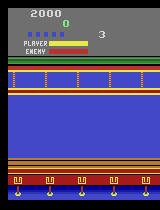}}
&
\textbf{KungFuMaster-ram} \hfill \textit{Atari (ALE)} \par\vspace{2pt}
\textbf{Task:} Fight through five floors of a temple to rescue a captive. The player uses punches and kicks to defeat waves of enemies---knife throwers, grippers, tom-toms, and snakes---while progressing rightward. Each floor ends with a boss fight. Cleared floors cycle with increasing difficulty. \par
\textbf{Obs:}~$\mathbb{R}^{128}$ (console RAM, \texttt{uint8}; semantically encodes player position, enemy positions and types, floor number, score). \par
\textbf{Action:} Discrete,~$|\mathcal{A}|=14$ (directions~$\times$ punch/kick combinations). \par
\textbf{Reward:}~$+100$--~$3000$ per enemy defeated (type-dependent),~$+2000$ per floor cleared (dense). \par
\textbf{Horizon:} 108{,}000 frames. Terminates when all lives are lost. \par
\textbf{Solved:} No standard threshold; higher scores reflect combat efficiency and floor progression.
\\ \midrule

\raisebox{-0.9\height}{\includegraphics[width=\linewidth]{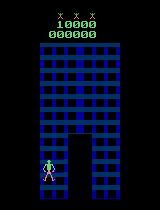}}
&
\textbf{CrazyClimber-ram} \hfill \textit{Atari (ALE)} \par\vspace{2pt}
\textbf{Task:} Scale the outside of four skyscrapers by grabbing windows with both hands while avoiding obstacles. Residents slam windows shut, a giant condor drops eggs, falling girders and signs must be dodged, and a King Kong--like gorilla blocks the path near the top. The player controls left and right hands independently. \par
\textbf{Obs:}~$\mathbb{R}^{128}$ (console RAM, \texttt{uint8}; semantically encodes player position, hand grip state, obstacle positions, building progress, score). \par
\textbf{Action:} Discrete,~$|\mathcal{A}|=9$ (8 directions + NOOP). \par
\textbf{Reward:} Points for vertical progress, bonus for reaching the top of each building (dense). \par
\textbf{Horizon:} 108{,}000 frames. Terminates when all lives are lost. \par
\textbf{Solved:} No standard threshold; higher scores reflect climbing speed and obstacle avoidance.
\\ \midrule

\raisebox{-0.9\height}{\includegraphics[width=\linewidth]{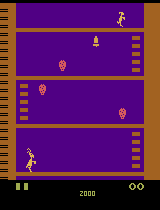}}
&
\textbf{Kangaroo-ram} \hfill \textit{Atari (ALE)} \par\vspace{2pt}
\textbf{Task:} A mother kangaroo must climb a multi-level structure to rescue her joey at the top. Monkeys throw apple cores from above, and the player punches them for points or dodges. Collecting fruit earns bonus points. A bell at the top must be rung to complete each level, which resets with increased difficulty. \par
\textbf{Obs:}~$\mathbb{R}^{128}$ (console RAM, \texttt{uint8}; semantically encodes player position, monkey positions, fruit locations, joey position, score). \par
\textbf{Action:} Discrete,~$|\mathcal{A}|=18$ (8 directions~$\times$ fire/no-fire + NOOP + fire-only). \par
\textbf{Reward:}~$+200$ per fruit,~$+100$ per monkey punched,~$+200$--~$1600$ for completing a level (dense). \par
\textbf{Horizon:} 108{,}000 frames. Terminates when all lives are lost. \par
\textbf{Solved:} No standard threshold; higher scores reflect efficient climbing and combat.
\\

\end{longtable}
\end{small}

\noindent
The four Atari environments use the console's 128-byte RAM as observation (~$\mathbb{R}^{128}$, \texttt{uint8}) instead of pixel frames, feeding directly into the same linear encoder--reservoir--decoder architecture without convolutional layers.
These RAM bytes encode semantically meaningful game state variables (player position, enemy positions, scores, etc.)\ that can be decoded via the OCAtari library~\citep{delfosse2024ocatari}, providing a pathway for extending the language game framework to arcade environments.

\section{Full Policy Similarity Results}
\label{sec:full-policy-similarity}

\begin{figure}[p]
    \centering
    \includegraphics[width=\textwidth,height=0.9\textheight,keepaspectratio]{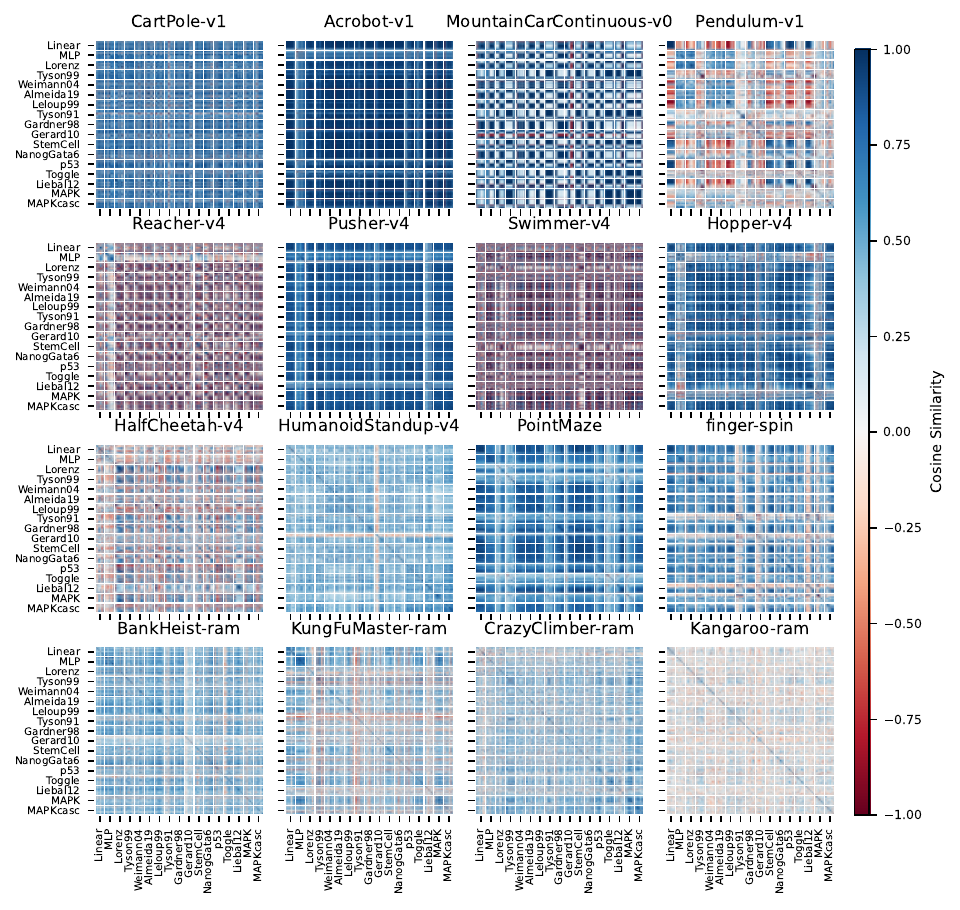}
    \caption{Policy similarity across all 17 reservoir architectures on all 16 RL environments. Each heatmap shows the cosine similarity between normalized action vectors of all trained policies (20 seeds~$\times$ 17 reservoirs = 340 policies per environment). Blue indicates strong positive correlation, white indicates no correlation, and red indicates negative correlation. White lines separate reservoir blocks. This extends Figure~\ref{fig:rational} in the main text to the complete environment suite.}
    \label{fig:rational-full}
\end{figure}

Figure~\ref{fig:rational-full} extends the policy similarity analysis from Section~\ref{sec:rational-agent-hypothesis} to all 16 reinforcement learning environments.

\paragraph{Similarity computation details.}
For continuous-action policies, the action vector is the policy mean. For discrete-action policies, we use the raw action logits after subtracting their within-state mean, so that task-irrelevant logit offsets do not affect cosine similarity. One-dimensional continuous-action environments require a small convention: a strictly one-dimensional vector would make cosine similarity depend only on sign, so we append a constant auxiliary coordinate of value~$0.1$ before normalizing. This is used only for the similarity diagnostic and does not affect the trained policies or reported rewards.

\paragraph{Low-similarity cells as a signature of under-training.}
The cells in Figure~\ref{fig:rational-full} with the weakest cross-architecture similarity coincide almost exactly with the environments in which training does not reach an accepted community benchmark. In Swimmer-v4 the community threshold is an average return of roughly~$360$, whereas our policies plateau at~$\sim 40$ across all reservoirs. HalfCheetah-v4 is considered well trained around~$4800$; most of our reservoirs sit around~$2000$ and the best-performing GRN only reaches~$\sim 3000$. Reacher-v4 has no official benchmark and accordingly yields a noisy similarity matrix. In contrast, tasks whose trained policies do reach their solved criterion display among the tightest cross-architecture similarity in the full panel: Acrobot-v1, for example, sits right at the community threshold of~$\geq -100$, and its block structure is correspondingly among the cleanest in Figure~\ref{fig:rational-full}. These are precisely the cells in which the Rational Agent Hypothesis is \emph{not expected} to hold on the under-trained side: convergence to a common behavioral class presupposes that the agents have actually converged to a rational optimum. Read this way, the pattern in Figure~\ref{fig:rational-full}---softened similarity exactly where policies fail to meet community benchmarks, and strong alignment wherever they do---is a consistency check on the hypothesis rather than a counter-example.

\section{Communication Pipeline Implementation Details}
\label{sec:communication-details}

This section provides implementation details for the three-step communication protocol described in Section~\ref{sec:talk-to-grn}.

\subsection{Environment Selection (Router Agent)}
\label{sec:router-agent}

The environment selection step uses a separate LLM call with structured output. Given the user's prompt, the LLM selects from the 16 trained environments by comparing the prompt's intent against a short textual description of each environment (2--3 sentences capturing the core objective and physical metaphor). The LLM's output is constrained via an enum schema so that only valid environment names can be returned.

For example, the prompt ``Keep the pole balanced'' is routed to CartPole-v1 (a stabilization task), while ``Run as fast as possible'' routes to HalfCheetah-v4 (a locomotion task), and ``Rescue the baby'' routes to Kangaroo-ram (a platforming and rescue task). Table~\ref{tab:router-examples} shows selected routing examples.

\begin{table}[h]
\centering
\small
\begin{tabular*}{\textwidth}{@{\extracolsep{\fill}}lll@{}}
\toprule
\textbf{Prompt} & \textbf{Selected Environment} & \textbf{Category} \\
\midrule
``Keep the pole balanced'' & CartPole-v1 & Classic Control \\
``Swing up and stabilize'' & Pendulum-v1 & Classic Control \\
``Run as fast as possible'' & HalfCheetah-v4 & MuJoCo Locomotion \\
``Navigate to the goal'' & PointMaze & Robotics \\
``Rob the vault and escape'' & BankHeist-ram & Atari RAM \\
``Rescue the baby'' & Kangaroo-ram & Atari RAM \\
``Kick the enemy'' & KungFuMaster-ram & Atari RAM \\
``Climb higher'' & CrazyClimber-ram & Atari RAM \\
\bottomrule
\end{tabular*}
\caption{Router Agent environment selection examples. The LLM matches the user's conversational intent to the most semantically appropriate RL environment.}
\label{tab:router-examples}
\end{table}

\subsection{Internal State Design and Rendering}
\label{sec:state-design}

In the human-to-system translation step, the LLM designs the environment's \emph{internal state} rather than the observation vector. The designed state is then set in the actual RL environment, which derives the observation that the policy was trained on. This ensures that the policy receives inputs in the exact format it expects, while the internal state can be rendered as a visual frame.

The state-setting mechanism varies by environment family:
\begin{itemize}
    \item \textbf{Classic Control} (CartPole, Pendulum, Acrobot, MountainCar): The internal state is a low-dimensional vector (e.g.,~$[\text{cart\_pos}, \text{cart\_vel}, \text{pole\_angle}, \text{pole\_vel}]$ for CartPole). The environment derives the observation from it (e.g., Pendulum converts~$[\theta, \dot\theta]$ to~$[\cos\theta, \sin\theta, \dot\theta]$).
    \item \textbf{MuJoCo} (Reacher, Pusher, Swimmer, Hopper, HalfCheetah, HumanoidStandup): The internal state is~$[\mathbf{q}_\text{pos}, \mathbf{q}_\text{vel}]$ (joint positions and velocities). The environment's \texttt{\_get\_obs()} method produces the observation (which may exclude root coordinates or include computed quantities).
    \item \textbf{Atari RAM} (BankHeist, KungFuMaster, CrazyClimber, Kangaroo): The 128-byte RAM serves as both internal state and observation. To produce valid rendered frames, we first warm up the emulator to a normal game state, then selectively overwrite only the documented RAM addresses (player position, enemy positions, score, lives, etc.) while preserving undocumented addresses that control rendering structures (maze layouts, floor patterns, sprite tables). The modified RAM is applied via the ALE \texttt{setRAM} interface, followed by a NOOP step to update the display.
    \item \textbf{dm\_control} (finger-spin): The internal state is the physics state vector (joint angles and velocities), set via \texttt{physics.set\_state()}.
    \item \textbf{Gymnasium Robotics} (PointMaze): The internal state is~$[x, y, v_x, v_y]$, set directly on the point mass's MuJoCo data.
\end{itemize}

\subsection{LLM System Prompts}
\label{sec:llm-prompts}

The communication pipeline uses four LLM system prompts across the three stages described in Section~\ref{sec:talk-to-grn}. All prompts use template variables \texttt{\{env\_name\}} (the selected environment name) and \texttt{\{env\_desc\}} (a detailed textual description of the environment loaded from documentation files). The router prompt additionally includes \texttt{\{env\_list\}}, a formatted list of all 16 trained environments with 2--3 sentence descriptions of each.

\begin{figure}[htb]
    \centering
    \small
    \begin{prompt}
You select the most appropriate RL environment to serve as the metaphorical frame for interpreting a user's conversational message.

Each environment defines a distinct vocabulary of states and actions. The chosen environment provides the context within which the user's intent is translated into the agent's behavior.

Available environments:
{env_list}

Given the user's message, choose the environment whose dynamics, goals, or metaphorical structure best match the user's intent, emotional state, or described situation. Return the exact env_name from the list above.

Think step by step: first identify the core concept in the user's message (e.g., balance, speed, fighting, navigation, reaching), then match it to the most fitting environment.
    \end{prompt}
    \caption{System prompt for the router agent, which selects the most semantically appropriate RL environment for a given user prompt. The \texttt{\{env\_list\}} placeholder is populated at runtime with the names and short descriptions of all 16 trained environments (see Table~\ref{tab:env-catalog}). The LLM's output is constrained via a structured enum schema so that only valid environment names can be returned.}
    \label{fig:system-prompt-router}
\end{figure}

\begin{figure}[htb]
    \centering
    \small
    \begin{prompt}
You are translate human natural language into a goal action for a rational agent in the {env_name} RL environment.

# Environment Description:
{env_desc}

# Requirement and Instruction

You will translate human natural language into a formal goal action for a rational agent in the {env_name} RL environment.

For example, in the CartPole-v1 environment, "Keep stable" can be translated into "Keep the pole upright". And "Move fast" can be translated into "Make the cart continuously move to left or right".

# Format

Directly output the translated goal state or action, without any other text.
    \end{prompt}
    \caption{System prompt for~$G(p_H)$, which infers the goal action from human natural language. Here \texttt{\{env\_name\}} is the name of the selected environment, and \texttt{\{env\_desc\}} is a detailed description of the environment.}
    \label{fig:system-prompt-goal}
\end{figure}

\begin{figure}[htb]
    \centering
    \small
    \begin{prompt}
Given {env_name} RL environment and a goal action, you design a environment state that lead a rational agent to take the goal action.
# Environment Description:
{env_desc}

# Requirement and Instruction

You will design a environment state that lead a rational agent to take the goal action.

For example, in the CartPole-v1 environment, if the goal action is "Move to left", you can design the environment state as:
(assume the vector represents [cart_position, cart_velocity, pole_angle, pole_velocity])
{{
    "state": [0.1, 0.0, -0.2, 0.0],
}}

# Format

Directly output the environment state in JSON format, with the key "state" and the value is a list of numbers.
    \end{prompt}
    \caption{System prompt for~$S_\text{LLM}$, which designs the environment state that leads a rational agent to take the goal action. Here \texttt{\{env\_name\}} is the name of the selected environment, and \texttt{\{env\_desc\}} is a detailed description of the environment.}
    \label{fig:system-prompt-env-design}
\end{figure}

\begin{figure}[htb]
    \centering
    \small
    \begin{prompt}
You are an agent in the {env_name} RL environment. The user will provide you the current state of the environment, your action, and a value-change signal Delta V (the PPO critic's value at the current state minus its value at a Gaussian reference observation). You will translate your action into natural language.

# Environment Description:
{env_desc}

# Instruction

Given the current state of the environment and the action, first infer the short-term goal of the action. Then use Delta V as an emotional signal: a clearly positive Delta V means the current state looks more promising than the reference (confident, hopeful tone); a clearly negative Delta V means it looks worse than the reference (uncertain, stressed tone); a near-zero Delta V means roughly indifferent (neutral, cautious tone). Finally, draft a reply to the user.

# Example (with CartPole-v1)

When the cart's pole is tilted to the left, the action is "Move to left", and Delta V is clearly negative, you can reply: "I'm not feeling good, and I'm trying to make the pole upright."

Note: Keep your reply concise and short in plain text. Do not directly talk about the numbers.

# Format

Directly output your reply in natural language, without any other text.
    \end{prompt}
    \caption{System prompt for~$\mathcal{T}_{S\to H}$, which translates the system's action into a natural language reply. Here \texttt{\{env\_name\}} is the name of the selected environment, and \texttt{\{env\_desc\}} is a detailed description of the environment.}
    \label{fig:system-prompt-reply}
\end{figure}

\end{document}